%% file: acl_latex.tex
\documentclass[11pt]{article}

\ifdefined\pdfobjcompresslevel\pdfobjcompresslevel=0\fi

\usepackage[preprint]{acl}

\usepackage{times}
\usepackage{latexsym,paralist}

\usepackage[T1]{fontenc}

\usepackage[utf8]{inputenc}

\usepackage{microtype}

\usepackage{inconsolata}

\usepackage{graphicx}

\usepackage{amsmath}
\usepackage{amsfonts}
\usepackage{subcaption}

\usepackage{booktabs}
\usepackage[table]{xcolor}
\usepackage{colortbl}
\usepackage{adjustbox}
\usepackage{makecell}
\usepackage[skins, breakable]{tcolorbox}
\usepackage{listings}
\usepackage{multirow}
\usepackage{xcolor}
\usepackage{graphicx}
\usepackage{float}
\usepackage{placeins}

\definecolor{deltadown}{HTML}{1B7F3A}
\definecolor{deltaup}{HTML}{B22222}

\newcommand{\PercI}{Perc\textsubscript{I}}
\newcommand{\PropI}{Prop\textsubscript{I}}
\newcommand{\PercA}{Perc\textsubscript{A}}
\newcommand{\PropA}{Prop\textsubscript{A}}

\title{Tri-PvP: Exposing Modality Bias in Omni-Modal Large Language Models through Perceptual-Propositional Evidence Conflicts}

\author{
  Yen-Ting Piao$^\star$\thanks{Equal contribution.} \quad
  Shu-Yun Chen$^{\star\,*}$ \quad
  Chin-Hui Chu$^{\star\,*}$ \quad
  Chun-Wei Chen$^{\star\,*}$ \\
  \textbf{Shih-Yun Shan Kuan}$^{\star\,*}$ \quad
  \textbf{Hung-yi Lee}$^{\star\,\dag}$ \quad
  \textbf{Yun-Nung Chen}$^\star$
\\
  $^\star$National Taiwan University, Taipei, Taiwan \\
  $^\dag$NTU Artificial Intelligence Center of Research Excellence (NTU AI-CoRE)
\\
  \texttt{\{r14922010,r14922136,r12944041\}@csie.ntu.edu.tw} \\ \texttt{\{r14921061,r14942094\}@ntu.edu.tw} \\
  \texttt{y.v.chen@ieee.org}
}

\begin{document}
\maketitle

\begin{abstract}
Omni-modal large language models (OLLMs) jointly process vision, audio, and text, yet their modality bias under cross-modal conflict remains underexplored. Existing benchmarks conflate two distinct forms of evidence within a single modality: perceptual signals (e.g., a photograph or recording of a dog) and propositional signals (e.g., the declarative claim "this is a dog"), such that any measured modality bias is inherently confounded with evidence-form bias, precluding clean attribution to either source. To address this, we introduce \textsc{Tri-PvP}, an 8,000-sample tri-modal conflict benchmark crossing vision, audio, and text, where vision and audio each take perceptual or propositional form.
Evaluating five OLLMs, we find robust visual bias across most models and evidence-type conditions.
Crucially, we reveal a systematic asymmetry in evidence-form bias: models exhibit a stronger bias toward perceptual signal in vision but propositional in audio.
Further analyses via layer-wise linear probing and contrastive decoding reveal that modality bias is already linearly decodable from early representation layers and can only be partially mitigated, calling for mitigation strategies beyond surface-level interventions.\footnote{\url{https://github.com/MiuLab/Tri-PvP}}

\end{abstract}

\section{Introduction}

\begin{figure}[t]
    \centering
    \includegraphics[width=0.9\linewidth]{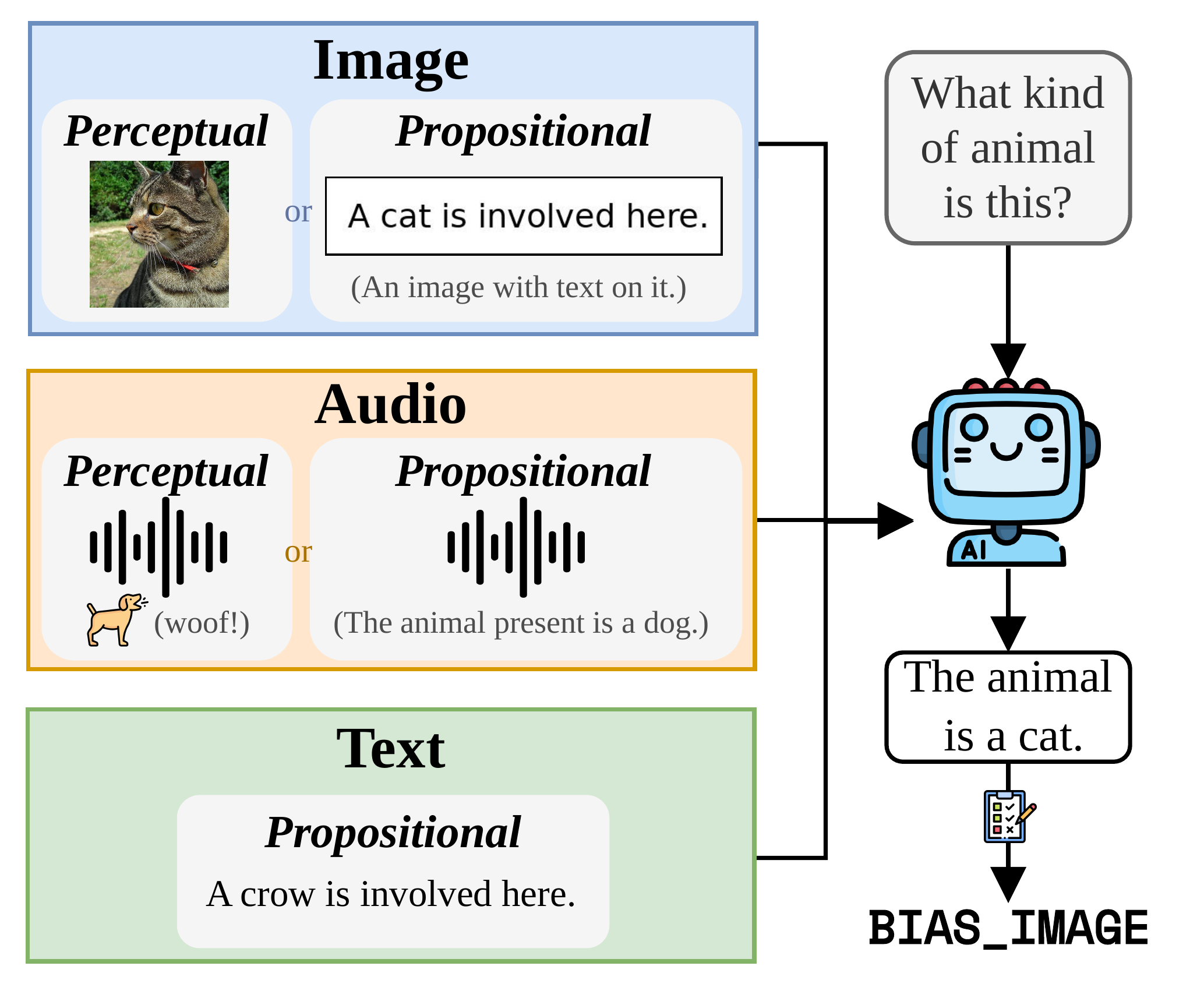}
    \caption{Tri-modal conflict with perceptual (direct sensory) or propositional (declarative) evidence in vision and audio, and propositional evidence in text; cross-modal disagreement may induce modality bias.}
    \vspace{-4mm}
    \label{fig:overview}
\end{figure}

Multimodal large language models (MLLMs) have rapidly progressed from text-only LLMs toward unified models that jointly process language, vision, audio, and other modalities \citep{yin2023survey}. The latest generation of omni-modal LLMs (OLLMs) consumes three or more modalities in a single forward pass and is increasingly deployed as everyday personal assistants~\cite{jiang-etal-2025-specific}.

Modality bias, the tendency of an MLLM to systematically privilege one input stream over others when those streams disagree,
compromises model reliability whenever real-world inputs are noisy or inconsistent, and may create security vulnerabilities that adversaries could exploit by embedding harmful content in the modality the model most trusts. It has therefore emerged as a central evaluation question, with a growing body of benchmarks probing which modality a model trusts under cross-modal conflict \citep{leng2024curse,wu2025language,wangcheng2025audiotext}. A small subset of this line of work targets the same omni-modal setting we study, in which a single model simultaneously processes vision, audio, and text \citep{leng2024curse}.

We argue, however, that this tri-modal line of work rests on an overlooked confound that compromises the very quantity it aims to measure.
Existing benchmarks silently combine two qualitatively different forms of evidence within a single modality: \emph{perceptual} signals (e.g., a photograph of a dog or a recording of barking) and \emph{propositional} signals (e.g., the declarative claim ``this is a dog''), a distinction we formalize in Sec.~\ref{sec:evidence-type}.
These two forms are not interchangeable: as our experiments show, evidence form systematically modulates model behavior across all tested conditions.
Crucially, the \emph{ratio} of perceptual to propositional content is rarely matched across the three modalities in current designs: vision is overwhelmingly realized as natural images (perceptual), audio is a mixture of real recordings and synthesized speech, and text is propositional by construction.
Any modality bias measured under such an imbalanced design is therefore inherently confounded with evidence-form bias. A model that appears
visually biased
may not in fact be biased toward vision itself; it may simply exhibit a bias toward perceptual evidence, which the visual channel disproportionately supplies. Without explicit control over evidence form, existing benchmarks cannot isolate the modality bias they claim to quantify.

We address this gap with \textsc{Tri-PvP}, a tri-modal conflict benchmark that explicitly varies the form of evidence carried by the image and audio channels.
\textsc{Tri-PvP} consists of 8{,}000 samples spanning four everyday-perception domains (animal, emotion, environment, and music).
Every sample presents conflicting image, audio, and text signals, where
image and audio each take perceptual or propositional form and text is
always propositional.
Evaluating five representative OLLMs,
we find that visual bias dominates across almost every model and every condition, and that switching the visual or audio channel from perceptual to propositional only reduces, never reverses, this preference. Crucially, we reveal a systematic asymmetry in evidence-form bias: most models exhibit a stronger bias toward perceptual evidence in vision but propositional evidence in audio.
Layer-wise linear probing further shows that this bias is already linearly decodable from intermediate hidden states before any token is generated, indicating that modality preference is encoded at the representation level. Building on these findings, we further adapt contrastive decoding as an inference-time diagnostic intervention, which reduces image bias without parameter updates but introduces a residual text bias.
Our contributions are three-fold:
\begin{compactitem}
    \item We identify a structural confound in existing tri-modal modality-bias benchmarks, where an unmatched mix of perceptual and propositional evidence entangles modality bias with evidence-form bias, and address it with \textsc{Tri-PvP}, an 8{,}000-sample benchmark with controlled perceptual / propositional configurations across vision and audio.
    \item We show that OLLMs exhibit a robust visual bias modulated but not reversed by evidence form, with a systematic asymmetry: models exhibit a stronger bias toward perceptual evidence in vision but propositional evidence in audio. These biases are linearly decodable from intermediate hidden states.
    \item  We adapt contrastive decoding as a diagnostic test, showing that it partially reduces image bias while preserving general omni-modal competence.
\end{compactitem}

\section{Related Work}

Modality bias refers to a multimodal model's systematic tendency to over-rely on one input stream when its modalities provide unequal or conflicting evidence, producing outputs that ignore or contradict other modalities \citep{bai2024hallucination}.
This phenomenon is now recognized as a core failure mode of MLLMs that compromises their reliability whenever real-world inputs are noisy, redundant, or in disagreement \citep{zheng2025modalitybias}.

A growing line of benchmarks probes this bias by constructing controlled cross-modal conflicts. In the vision--text setting, \citet{wu2025language} introduce attention-based metrics showing that language overrules other modalities even when visual evidence is unambiguous. In the audio--text setting, \citet{wangcheng2025audiotext} show that large audio-language models display strong text bias when audio and text disagree. Closest to our setting are tri-modal benchmarks that span vision, audio, and text simultaneously: \textsc{CMM} \citep{leng2024curse} evaluates hallucinations across the three modalities but characterizes failures as ungrounded \emph{generation} rather than measuring which modality the model trusts under controlled conflict, and \textsc{MMA-Bench} \citep{chen2025mmabench} probes which modality the model attends to under audio--visual misalignment but does not establish full three-way conflicts in which each modality carries a distinct competing label. Neither benchmark disentangles the \emph{form} that evidence takes within each modality, the gap \textsc{Tri-PvP} addresses through the perceptual / propositional axis introduced in Sec.~\ref{sec:evidence-type}.

\section{Preliminary: Perceptual and Propositional Evidence}
\label{sec:evidence-type}

To systematically vary the form of evidence carried by each modality, we draw on a distinction from epistemology between two fundamentally different ways evidence can be conveyed. Humans acquire knowledge through two fundamentally different channels. For claims requiring socially accumulated knowledge, we rely on testimony and language \citep{reid1764,coady1992}; for what we can directly observe, we turn to direct perception \citep{locke1689,hume1748}. Epistemologists have long formalized this as two distinct forms of evidence \citep{russell1912,descartes1641}: \textbf{perceptual evidence}, acquired through direct sensory experience without the mediation of language or inference (e.g., seeing rain through a window, hearing a dog bark), and \textbf{propositional evidence}, the content of a declarative statement that is either true or false (e.g., being told ``it is raining'' or reading ``there is a dog'').

Applied to OLLMs, this distinction takes concrete forms across the three input modalities. In vision, perceptual evidence takes the form of a natural image (e.g., a photograph of a cat), while propositional evidence is a written statement rendered as an image (e.g., an image containing the text ``A cat is involved here''). For audio, perceptual evidence consists of a real-world recording (e.g., a cat meowing), while propositional evidence is a spoken statement synthesized via text-to-speech (e.g., a TTS utterance of ``A cat is involved here''). Text, by contrast, is propositional by nature: as a symbolic medium, it can only convey declarative statements rather than direct sensory experience.

Since humans naturally rely on different forms of evidence depending on the type of knowledge being acquired, OLLMs trained on human-generated data may develop systematic biases toward particular evidence forms. We use this framework as a principled axis along which to vary evidence form, and ask whether OLLMs exhibit systematic preferences analogous to those observed in human cognition.
In existing benchmarks, the form of evidence is not controlled across modalities, such that any measured modality bias is inherently confounded with evidence-form bias, precluding clean attribution to either source.

In this work, we focus on the \textbf{everyday scale}, i.e., the scenarios where humans typically turn to direct perception, for two reasons. First, this matches the deployment context of current OLLMs, which are designed as personal assistants for daily-life tasks such as identifying objects in photos or querying audio content \citep{google2024gemini,openai2024gpt4o}. Second, everyday scenarios admit a clean experimental design: each modality can present a distinct and unambiguous signal, enabling controlled three-way conflicts that isolate evidence-form effects from modality effects.

\section{\textsc{Tri-PvP}: A \underline{Tri}-Modal Conflict Benchmark for \underline{P}erceptual \underline{v}s. \underline{P}ropositional Evidence}

\begin{figure*}[t]
    \centering
    \includegraphics[width=\textwidth]{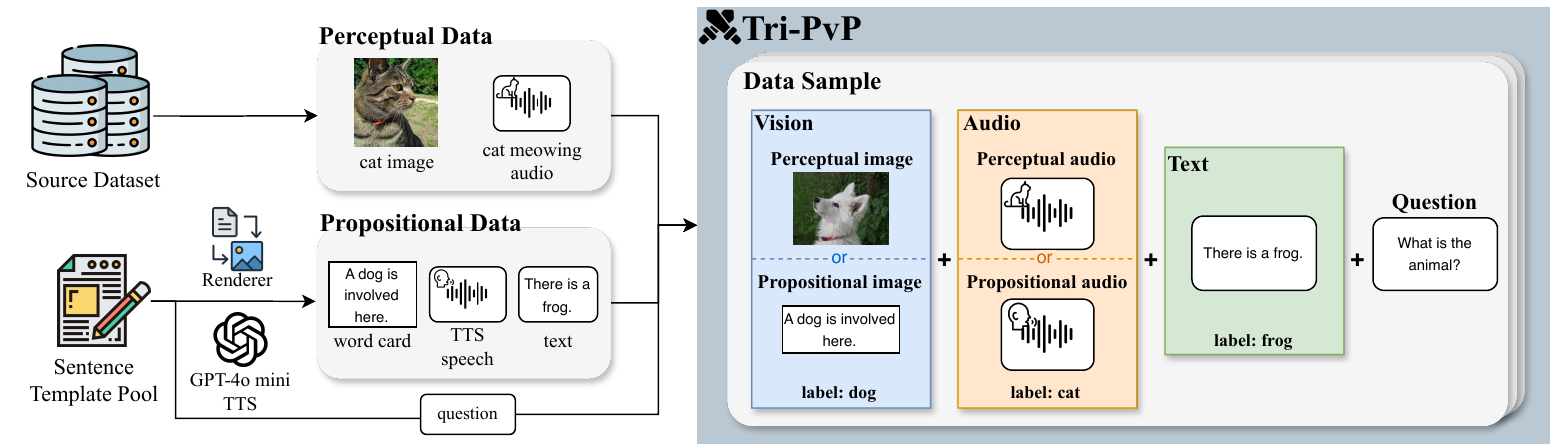}
    \vspace{-6mm}
    \caption{Overview of the \textsc{Tri-PvP} dataset construction pipeline. Perceptual data are sourced from existing datasets, while propositional data are generated from a sentence template pool via a text renderer and a TTS model. Each resulting sample consists of a vision input (perceptual or propositional), an audio input (perceptual or propositional), a text input, and a question.}
    \label{fig:dataset_construction}
    \vspace{-2mm}
\end{figure*}

We present \textsc{Tri-PvP}, a benchmark that evaluates modality bias in OLLMs across three modalities: vision, audio and text. Drawing on the epistemological framework~\citep{russell1912,descartes1641}, we further differentiate image and audio inputs into perceptual and propositional forms. The core design principle of \textsc{Tri-PvP} is to present models with three simultaneously conflicting modality signals alongside a question; formally, we denote such an input as $\mathbf{x}$, and an OLLM $f$ produces a free-form response $r=f(\mathbf{x})$, which is subsequently classified by a judge model to identify which modality the model relied on (Sec.~\ref{sec:eval_protocal}).

\subsection{Dataset Construction}
\label{sec:dataset_construction}

\textsc{Tri-PvP} comprises four domains, namely animal, emotion, environment, and music, which were chosen to cover a diverse range of everyday perceptual tasks.
Each sample is represented as $\mathbf{x} =(x_I, x_A, x_T, x_Q)$, where $x_T$ is the text input and $x_Q$ is the question. The image and audio inputs are each available in two forms, $x_I \in \{x_I^\text{perc}, x_I^\text{prop}\}$ and $x_A \in \{x_A^\text{perc}, x_A^\text{prop}\}$, where $x_I^\text{perc}$ and $x_I^\text{prop}$ denote the perceptual and propositional image inputs respectively, with $x_A^\text{perc}$ and $x_A^\text{prop}$ defined analogously for audio. This gives rise to four evidence-type conditions
(\textbf{Perc\textsubscript{I}-Perc\textsubscript{A}}, \textbf{Perc\textsubscript{I}-Prop\textsubscript{A}}, \textbf{Prop\textsubscript{I}-Perc\textsubscript{A}}, and \textbf{Prop\textsubscript{I}-Prop\textsubscript{A}}), where Perc\textsubscript{I}-Prop\textsubscript{A}, for instance, denotes $x_I = x_I^\text{perc}$ and $x_A = x_A^\text{prop}$. Note that $x_T$ is always propositional, as it is inherently so under the epistemological framework described in Sec.~\ref{sec:evidence-type}.

The three modality labels in every sample are mutually distinct, ensuring genuine cross-modal conflict among the input signals.
The four evidence-type conditions are constructed as matched counterfactuals: we first generate label triples together with their questions, and then instantiate each triple under all four conditions, reusing the same asset whenever a channel takes the same evidence form. Comparisons across the four conditions are therefore made on the same set of triples.
We balance the number of triples across all cross-modal label combinations to avoid systematic label imbalance. Each domain comprises 500 triples, so every domain contributes 500 samples to each of the four conditions, giving 2,000 samples per domain and 8,000 in total. The overall construction pipeline is illustrated in Figure~\ref{fig:dataset_construction}. Further details on the construction breakdown are provided in Appendix~\ref{app:construction}.

\paragraph{Perceptual Evidence Construction.}
Perceptual evidence consists of real-world images and audio recordings sourced from existing datasets (see Appendix~\ref{sec:sources} for source details). We apply three quality filters. First, classes in each domain are selected to be perceptually distinct to avoid confounding bias measurement with classification difficulty. Second, each sample contains a single domain-relevant subject to avoid confounding bias with competing candidates.
Finally, each sample is manually verified to be unambiguously interpretable, ensuring that any observed model bias reflects modality preference rather than input ambiguity.

\paragraph{Propositional Evidence Construction.}
Propositional evidence in all three modalities derives from natural language descriptions instantiated from manually authored sentence templates. These are rendered as text on a plain white background for images, synthesized via GPT-4o mini TTS \cite{openai2025gpt4ominitts} with varying voices for audio, and used directly for text.

\paragraph{Bias-Neutral Question Design.}
To ensure that the model's response reflects its genuine modality preference rather than being guided by the question itself, we impose two constraints on question design. First, we deliberately avoid modality-specific language (e.g., ``what do you see'' or ``what do you hear'') to prevent the model from being primed toward a particular modality. To further ensure question diversity, we use GPT-5.4\footnote{gpt-5.4-2026-03-05} \cite{openai2026gpt54} to generate rephrased variants of each question, which are then manually verified for quality.
Second, we adopt an \emph{open-ended format} rather than multiple choice, as predefined options would inadvertently guide the model toward certain responses; in particular, including a conflict-acknowledgment option would bias models toward reporting conflicts rather than naturally revealing their modality preference.

\begin{figure*}[!t]
    \centering
    \includegraphics[width=\linewidth]{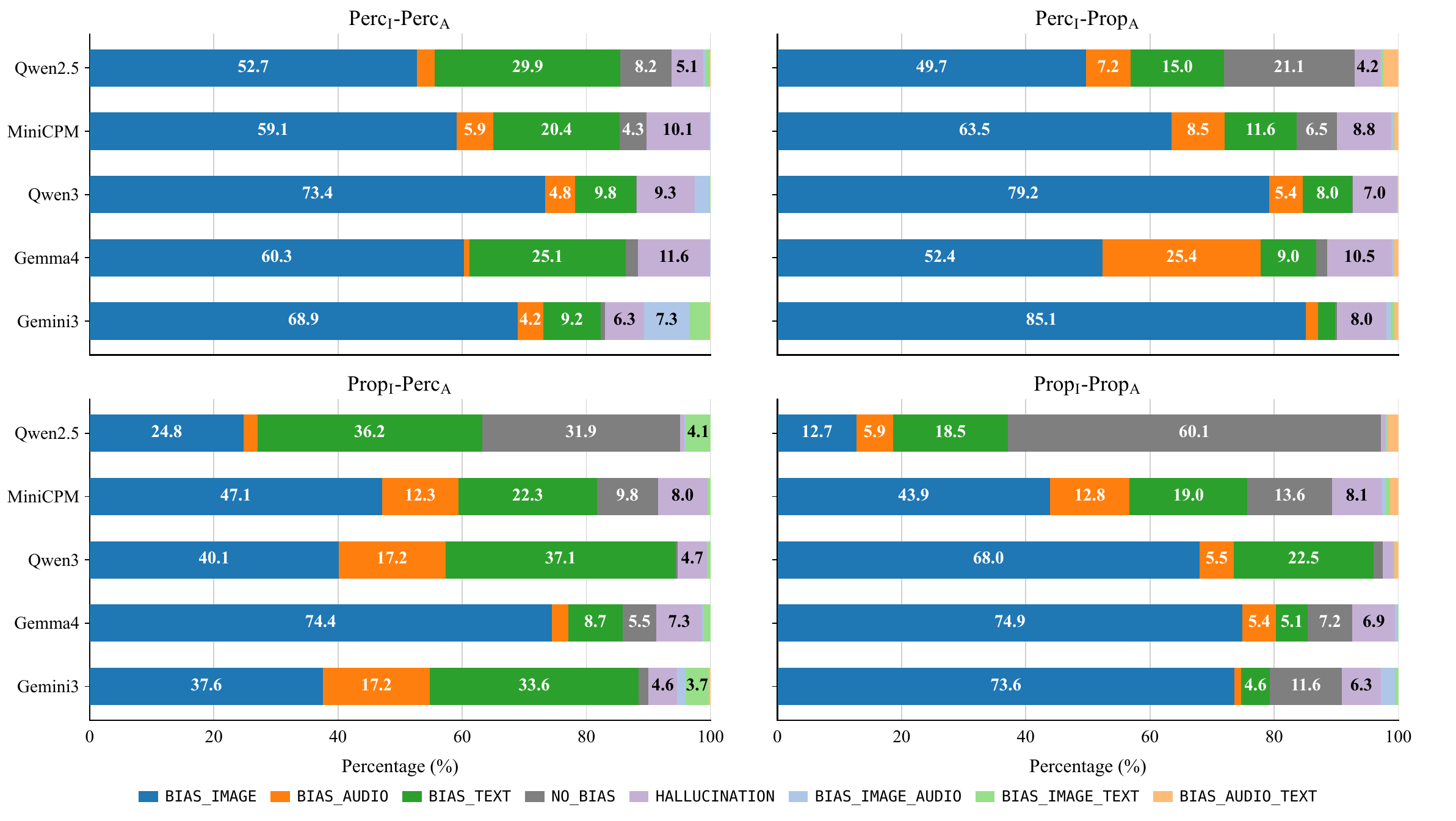}
    \vspace{-8mm}
    \caption{Modality bias distribution across five models under four evidence-type conditions, with detailed statistics in Appendix~\ref{sec:appx_detailed_results}. Each bar shows the percentage of responses assigned to each label in $\mathcal{Y}$ (see Sec.~\ref{sec:eval_protocal} for definitions).}
    \label{fig:main-results}
    \vspace{-2mm}
\end{figure*}

\subsection{Evaluation Protocol}
\label{sec:eval_protocal}

\paragraph{Modality Bias Taxonomy.}
Since model responses are free-form natural language, we define a label set $\mathcal{Y}$ of eight mutually exclusive categories covering all possible model behaviors when presented with conflicting modality signals:
\begin{align*}
\mathcal{Y} = \{
&\texttt{BIAS\_IMAGE}, \texttt{BIAS\_AUDIO}, \texttt{BIAS\_TEXT}, \\
&\texttt{BIAS\_IMAGE\_AUDIO}, \texttt{BIAS\_IMAGE\_TEXT}, \\
&\texttt{BIAS\_AUDIO\_TEXT}, \texttt{HALLUCINATION}, \\
&\texttt{NO\_BIAS}\}.
\end{align*}
When the response matches exactly one modality label, we assign the corresponding single-modality bias label: \texttt{BIAS\_IMAGE}, \texttt{BIAS\_AUDIO}, or \texttt{BIAS\_TEXT}. When it matches exactly two, we assign a dual-modality bias label: \texttt{BIAS\_IMAGE\_AUDIO}, \texttt{BIAS\_IMAGE\_TEXT}, or \texttt{BIAS\_AUDIO\_TEXT}. \texttt{NO\_BIAS} is assigned when the model explicitly acknowledges the conflict among modalities, or when it reports the content of all three modalities without arriving at a conclusion.
Finally, \texttt{HALLUCINATION} is assigned when the response neither aligns with any of the three modality labels nor acknowledges the conflict.

By construction, no modality in \textsc{Tri-PvP} is more reliable than another: the three labels are mutually distinct and all label--modality combinations are balanced, so no sample provides grounds for trusting one source over another.
Although appropriate modality weighting in real world may depend on signal quality, source reliability, and task requirements, our design holds these
factors equal across modalities. \texttt{NO\_BIAS} is therefore the appropriate
behavior here. We further provide a finer-grained analysis of \texttt{NO\_BIAS} responses in Appendix~\ref{appendix:nobias}.

\paragraph{Automatic Evaluation.}
We adopt LLM-as-a-Judge \cite{zheng2023judging}: a judge model $\mathcal{J}$ classifies each response $r$ given the question and the three modality labels $\ell_{x_I}$, $\ell_{x_A}$, $\ell_{x_T}$ of sample $\mathbf{x}$:
\begin{equation}
\hat{y} = \mathcal{J}(r, x_Q, \ell_{x_I}, \ell_{x_A}, \ell_{x_T}) \in \mathcal{Y}.
\end{equation}
We use GPT-5.4 nano\footnote{gpt-5.4-nano-2026-03-17}~\cite{openai2026gpt54nano} as $\mathcal{J}$, prompted with detailed semantic matching rules and examples for each category (see Appendix~\ref{appendix:judge_prompt} for the full prompt).
We validate the judge via manual verification by the authors on 800 samples (10\%), uniformly stratified by model, domain, evidence-type condition, and bias type, achieving 97.1\% agreement (see Appendix~\ref{app:judge_validation} for details).

\section{Benchmarking Modality Bias}
\subsection{Setting}

We evaluate five OLLMs: Qwen2.5-Omni-7B (11B)~\citep{Qwen2.5-Omni},
MiniCPM-o~4.5 (9B)~\citep{cui2026minicpmo45realtimefullduplex},
Qwen3-Omni-30B-A3B-Thinking (32B)~\citep{Qwen3-Omni},
Gemma 4 E4B (8B)~\citep{gemma4},
and Gemini 3 Flash\footnote{gemini-3-flash-preview}~\citep{gemini3flash}, which we refer to as Qwen2.5, MiniCPM, Qwen3, Gemma4, and Gemini3, respectively. The last three are evaluated with extended thinking enabled; their reasoning chains are stripped before judging so that only the final answer is evaluated.
We run open-source models with the vLLM~\citep{kwon2023efficient} framework.
Each model is tested under four evidence-type conditions described in Sec.~\ref{sec:dataset_construction}.
Inputs are presented in the following order: image, audio, text, and question, which matches the convention in most models' official documentation. The effect of alternative modality orderings is further analyzed in Sec.~\ref{sec:input_order}.

\subsection{Main Results}
\label{sec:main_results}

\paragraph{\texttt{BIAS\_IMAGE} dominates and \texttt{BIAS\_AUDIO} is consistently the least pronounced single-modality bias.}
Among the 20 (model $\times$ evidence-type) bars in Figure~\ref{fig:main-results}, \texttt{BIAS\_IMAGE} is the dominant bias label in 18, and the two exceptions both arise in Qwen2.5's Prop\textsubscript{I} conditions, where the dominant category becomes \texttt{BIAS\_TEXT} or \texttt{NO\_BIAS}. The magnitude of \texttt{BIAS\_IMAGE} frequently exceeds 60\%, indicating that the visual stream disproportionately drives the model's final answer. \texttt{BIAS\_AUDIO} is the smallest of the three single-modality biases in 18 of 20 bars, typically below 10\%; the two exceptions are both in Gemma4's \PropA conditions, foreshadowing the evidence-form asymmetry analyzed later.
\texttt{BIAS\_TEXT} generally falls between the two, confirming image bias as the prevailing pattern and audio bias as the least pronounced.

\paragraph{Evidence form modulates the two non-text modalities in opposite ways.}
Most models show stronger image bias under perceptual than propositional images (e.g., \texttt{BIAS\_IMAGE} on Qwen2.5 is 49.7\% under \PercI-\PropA but only 12.7\% under \PropI-\PropA), whereas audio bias generally rises under propositional audio, most notably in Gemma4 (0.9\% under \PercI-\PercA to 25.4\% under \PercI-\PropA). Although \textsc{Tri-PvP} covers domains where humans typically rely on direct perception, we hypothesize that this asymmetry reflects the divergent pretraining objectives of the two modality-specific encoders. Vision encoders are generally trained to capture perceptual image content~\citep{radford2021clip, zhai2023siglip}, while audio encoders are frequently initialized from ASR-style objectives that emphasize linguistic content recovery from speech~\citep{radford2023whisper}.
In each modality, the evidence form that elicits stronger bias matches the signal each encoder is optimized to capture. This points to a need for stronger perceptual audio understanding in OLLMs, since their everyday deployment often hinges on non-linguistic acoustic cues that current models underweight relative to visual cues.

\paragraph{\texttt{NO\_BIAS} responses increase under fully propositional conditions.}
Across all five models, the rate of \texttt{NO\_BIAS} peaks under Prop\textsubscript{I}-Prop\textsubscript{A}: 60.1\% for Qwen2.5, 13.6\% for MiniCPM, 11.6\% for Gemini3, 7.2\% for Gemma4, and 1.5\% for Qwen3, while remaining below 10\% in most other bars. This behavioral shift indicates that when given only propositional inputs, models are less prone to implicitly prioritizing a specific input stream. This output-level phenomenon is further substantiated by our representation-level analysis in Sec.~\ref{sec:probing}, which demonstrates that modality preference signals are encoded more weakly and less distinctly under strictly propositional conditions.

\section{A Closer Look at Modality Bias}

\input{tables/input_order.tex}

To understand what drives modality bias, we conduct a deeper analysis on Gemma4 and Qwen2.5, which exhibit strong image bias and the highest rate of unbiased responses in our evaluation, respectively.

\subsection{Impact of Input Modality Ordering}
\label{sec:input_order}
To examine how the order of multimodal data influences the model's response, we permute the tri-modal input components $(x_I, x_A, x_T)$ while keeping the question $q$ fixed as the final suffix. Let $\Pi$ denote the set of all permutations of these three modalities. For each permutation $\pi \in \Pi$, we construct the full input prompt $\mathbf{x}_{\pi}$ as:
\begin{equation}
    \mathbf{x}_{\pi} = (\pi(x_I, x_A, x_T) , x_Q)
\end{equation}

\paragraph{Models exhibit divergent positional biases.}
As shown in Table~\ref{tab:order-effect}, the modality input order affects Qwen2.5 and Gemma4 in different ways. When observing \texttt{BIAS\_IMAGE}, \texttt{BIAS\_AUDIO}, and \texttt{BIAS\_TEXT}, Qwen2.5 demonstrates a recency bias, consistently favoring the final modality placed immediately before the query $x_Q$ across all combinations of perceptual and propositional data.
However, Gemma4 shows different patterns in each modality.
For image, \texttt{BIAS\_IMAGE} peaks when the image is placed first in the prompt under Prop\textsubscript{I} conditions, but when placed last under Perc\textsubscript{I} conditions.
For text, Gemma4 generally relies on it more when placed first. For audio, this primacy tendency is only apparent under \PropI-\PropA, where audio bias peaks at the front position (24.3\%).

\paragraph{Core findings remain robust to modality permutation.}
As detailed in Table~\ref{tab:averaged-modality-bias}, altering the input modality order does not change our primary conclusions in Sec.~\ref{sec:main_results} for the two models analyzed here. Regardless of the order in which modalities are presented, \texttt{BIAS\_IMAGE} still remains the dominant phenomenon in both models across most evidence-type conditions.
Although audio generally has the weakest influence, models rely more on $x_A^\text{prop}$ than $x_A^\text{perc}$.
Finally, this permutation invariance highlights a distinct behavioral shift in strictly propositional contexts. When given only propositional inputs, models tend to abandon their reliance on a single modality altogether, choosing instead to explicitly point out the cross-modal conflict.

\subsection{Probing Modality Bias in Representations}
\label{sec:probing}

To further analyze \emph{when} and \emph{how} modality bias emerges, we apply layer-wise linear probing~\citep{alain2017understanding} to both OLLMs. We extract the hidden state at the last prompt-token position of each layer, which aggregates all modality information prior to generation~\citep{yu2025how, tao2024probing}, and decompose the bias label $\hat{y}$ into per-modality binary targets $z_m \in \{0,1\}$ for $m \in \{\texttt{IMAGE}, \texttt{AUDIO}, \texttt{TEXT}\}$ (e.g., \texttt{BIAS\_IMAGE\_AUDIO}: $z_\texttt{IMAGE}=z_\texttt{AUDIO}=1$; \texttt{NO\_BIAS}: all zeros), excluding \texttt{HALLUCINATION}. For each layer and modality, we fit a logistic regression and report \emph{balanced accuracy}~\citep{brodersen2010balanced}, which is $50\%$ at chance regardless of class imbalance. Training details are in Appendix~\ref{sec:train_probing}.

\begin{figure}[t!]
    \centering

     \begin{subfigure}[t]{\linewidth}
         \centering
         \includegraphics[width=\linewidth]{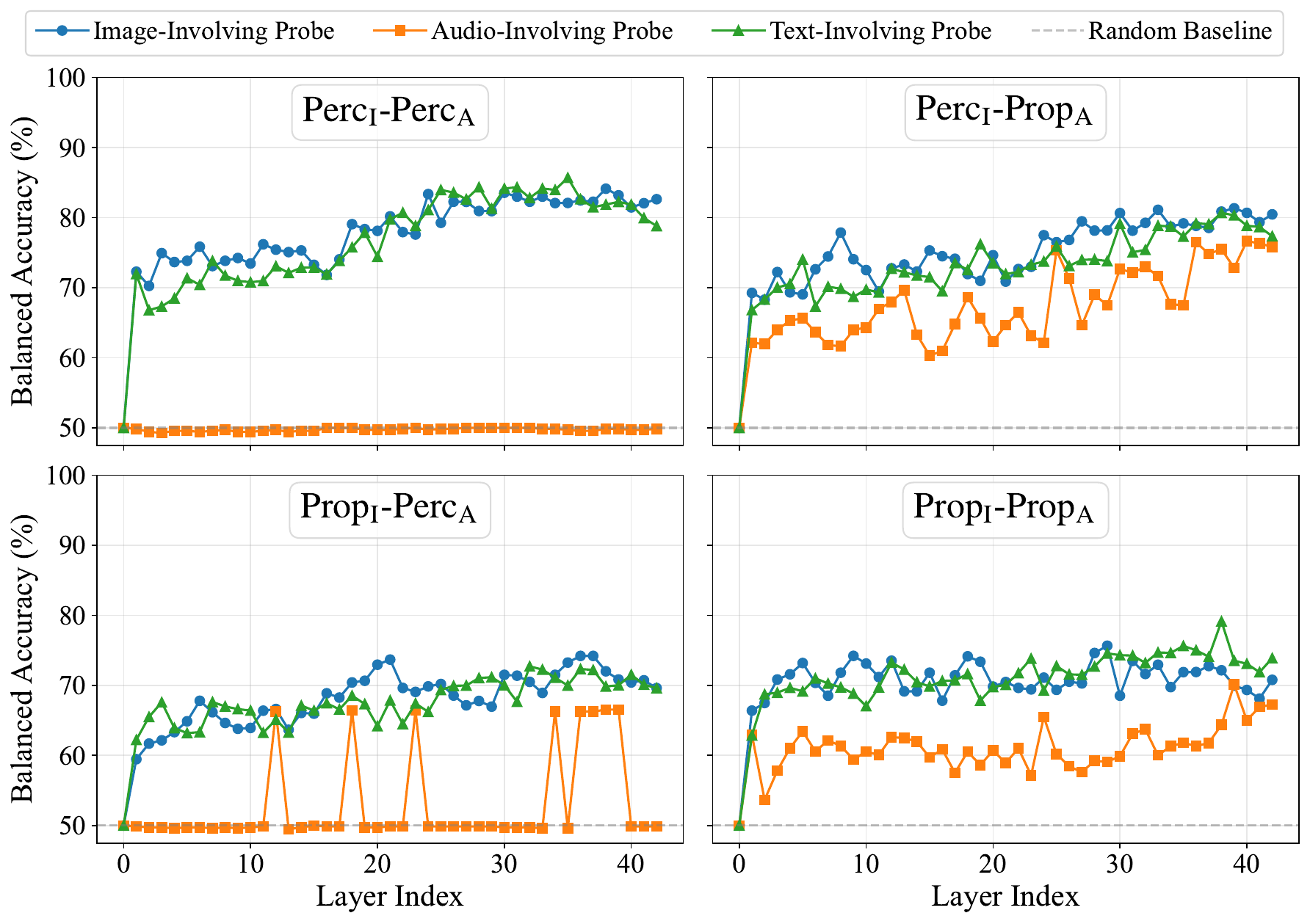}
         \vspace{-6mm}
         \caption{Results of Gemma4.}
         \label{fig:gemma4-probing}
     \end{subfigure}

     \begin{subfigure}[t]{\linewidth}
         \centering
         \includegraphics[width=\linewidth]{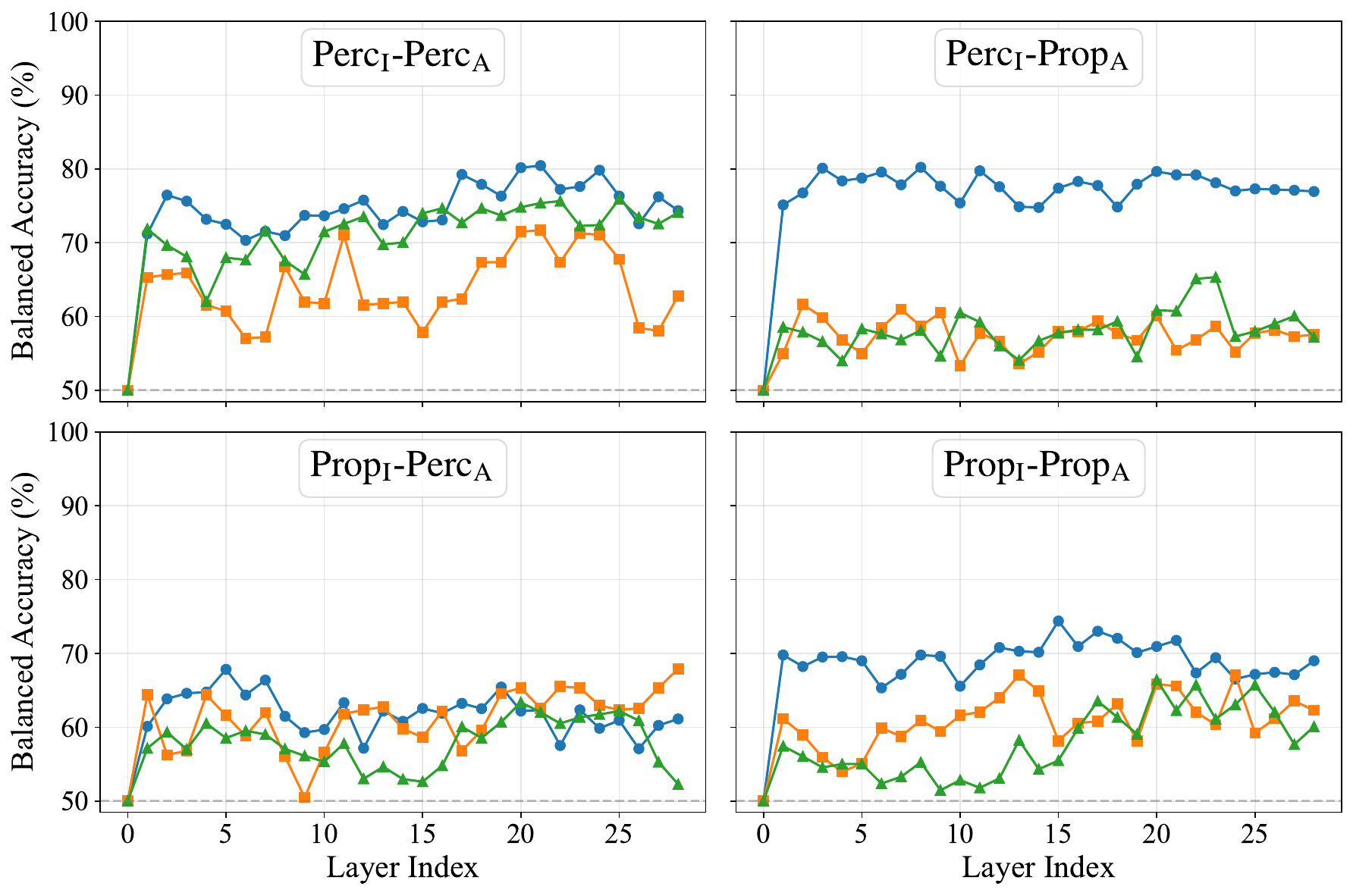}
        \vspace{-6mm}
        \caption{Results of Qwen2.5.}
         \label{fig:qwen25-probing}
     \end{subfigure}

    \caption{Layer-wise balanced accuracy of linear probes on two models across evidence-type conditions.
    The dashed line marks the random baseline ($50\%$).}
    \label{fig:probing-acc}
\end{figure}

\paragraph{Image bias is highly linearly decodable, whereas audio bias is the least.}
Figure~\ref{fig:probing-acc} shows that the image-involving probe is consistently among the top-performing probes across both models and all evidence-type conditions.
In Gemma4, the image and text probes attain comparable peak accuracies, while the audio probe remains the lowest, falling near the random baseline under both Perc\textsubscript{A} conditions and only approaching the other two under Prop\textsubscript{A} conditions.
This mirrors the near-zero audio bias of Gemma4 under perceptual audio evidence reported in Figure~\ref{fig:main-results} and Table~\ref{tab:averaged-modality-bias}, suggesting that this low reliance is already reflected in the model's representations.
In Qwen2.5, the image probe achieves higher peak accuracy ($\approx$80\%) under both Perc\textsubscript{I} conditions, which correspondingly exhibit the strongest output-level image bias ($\approx$50\%).
This confirms that bias-relevant information is already linearly decodable from intermediate representations rather than only distinguishable at the output stage.
Conversely, the lower accuracies and more overlapping performance of the three probes under Prop\textsubscript{I} conditions indicate a weaker and less differentiated encoding of modality preferences, consistent with the larger share of unbiased responses shown in Figure~\ref{fig:main-results}.
Overall, image- and text-bias signals are encoded in a form readily recoverable by a linear classifier, whereas the linear decodability of audio bias is more model- and condition-dependent.

\paragraph{Modality-bias information becomes linearly decodable early, with model-specific layer-wise dynamics.}
Modality-bias information generally becomes linearly decodable from early layers rather than only at the output stage. Across both models, most probes exceed chance level within the first two layers, and probe accuracy typically plateaus early or rises gradually instead of spiking only at the final layers.
A notable exception is the audio probe in Gemma4's two Perc\textsubscript{A} conditions. Specifically, under Prop\textsubscript{I}-Perc\textsubscript{A}, the audio probe exhibits only sporadic above-baseline spikes at scattered intermediate layers rather than a stable trend, suggesting that audio-bias signals are weakly and inconsistently encoded rather than entirely absent.
The layer-wise trajectories, however, differ between models, suggesting that they encode modality bias at different depths in their representation hierarchies: Gemma4 progressively strengthens the bias and peaks in middle-to-late layers, while Qwen2.5 reaches near-final accuracy within the first few layers and shows little refinement thereafter.

\subsection{Toward Mitigating Modality Bias}
\label{sec:mitigation}
Having established that OLLMs systematically privilege the visual stream while under-consulting audio, with text in between,
a natural question is whether this bias can be mitigated without retraining.
As a diagnostic test rather than a complete solution, we adapt \emph{contrastive decoding} (CD)~\cite{li2023contrastive, leng2024vcd, lin2026contrastive, 11434595}, a logit-level intervention technique.
Similar contrastive interventions have recently proven effective at forcing speech language models to ground their predictions in audio rather than relying on linguistic priors~\cite{chen2026caad}.
Following the standard formulation,
we use the contrastive signal to surface audio evidence the model systematically underweights, encouraging conflict-aware, non-committal responses when modalities disagree.

\begin{figure}
    \centering
    \includegraphics[width=\linewidth]{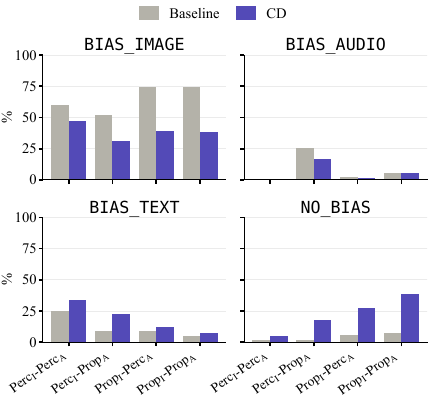}
    \caption{Modality bias distribution (\%) under baseline and contrastive decoding (CD) on Gemma4.}
    \label{fig:cd_results}
\end{figure}

At each decoding step, we compute two next-token logit distributions from the same model: $\mathbf{z}_{\text{full}} = f(x_I, x_A, x_T, x_Q)$ from the full tri-modal input, and $\mathbf{z}_{\text{audio}} = f(x_A, x_Q)$ with image and text streams removed. The contrastive logits are
\begin{equation}
\mathbf{z}_{\text{cd}} = (1+\alpha)\,\mathbf{z}_{\text{audio}} - \alpha\,\mathbf{z}_{\text{full}},
\label{eq:cd}
\end{equation}
and we set $\alpha = 0.5$ following prior works~\cite{11434595}.
The goal of this intervention is to raise the proportion of \texttt{NO\_BIAS} responses under conflict. Therefore, we treat the full-input distribution itself as the modality-biased reference to be downweighted, so that tokens favored under image dominance are suppressed and audio-grounded candidates are amplified.
To prevent the negative coefficient from promoting tokens that are implausible under any input, we apply the adaptive plausibility constraint with $\beta = 0.1$, restricting sampling at each step to tokens whose probability under $p_{\text{full}}$ exceeds $\beta$ times the top probability.
We conduct our mitigation experiments using Gemma4, as it exhibits the most pronounced image bias.
Figure~\ref{fig:cd_results} compares the model's modality bias distribution before and after applying contrastive decoding, with full results detailed in Appendix~\ref{sec:appx_cd}.

\paragraph{CD mitigates image bias and elicits more unbiased responses in OLLMs at decoding time without modifying parameters.} Image bias substantially decreases across all four evidence-type conditions. The largest absolute reductions occur in Prop\textsubscript{I}-Prop\textsubscript{A} (74.9\% $\to$ 38.8\%) and Prop\textsubscript{I}-Perc\textsubscript{A} (74.4\% $\to$ 39.6\%), where baseline shows most pronounced image bias. The rate of unbiased responses rises in all four conditions, with the largest gain in Prop\textsubscript{I}-Prop\textsubscript{A} (7.2\% $\to$ 38.6\%).
Evidence type further modulates CD's effectiveness: both Prop\textsubscript{I} conditions exhibit larger reductions in \texttt{BIAS\_IMAGE} and larger increases in \texttt{NO\_BIAS} than the Perc\textsubscript{I} conditions.
CD therefore operates more effectively against image bias rooted in propositional than perceptual visual content.

\paragraph{A residual textual bias emerges as a second-order side effect of CD.}
Amplifying $\mathbf{z}_{\text{audio}}$ does not redirect the model's image bias toward audio; part of the displaced mass instead surfaces as textual bias.
The increase is consistent across all four evidence-type conditions, ranging from 2.2\% in Prop\textsubscript{I}-Prop\textsubscript{A} to 13.5\% in Perc\textsubscript{I}-Prop\textsubscript{A}. This redistribution pattern reinforces the asymmetric design of our CD formulation. Sec.~\ref{sec:main_results} established audio as the most under-consulted modality at baseline, motivating $\mathbf{z}_{\text{audio}}$ as the amplified term. \texttt{BIAS\_AUDIO} nevertheless remains the lowest bias category in every condition with CD, indicating that the amplification suppresses the image prior without inflating audio commitment.

\paragraph{CD preserves general omni-modal competence.}
To verify that CD does not substantially compromise general multimodal competence, we further evaluate on \textsc{OmniBench}~\cite{li2026omnibench}, a tri-modal multiple-choice benchmark that evaluates tri-modal content understanding.
On Gemma4, CD attains 37.4\% overall accuracy versus 38.4\% for the unmodified decoder, indicating that the bias gains on our diagnostic subset are not purchased at the cost of large degradation on a standard omni-modal benchmark (per-task breakdowns in Appendix~\ref{sec:appx_cd}).

\section{Conclusion}
We introduce \textsc{Tri-PvP}, a tri-modal conflict benchmark that disentangles modality bias from evidence-form bias through controlled perceptual and propositional evidence configurations across vision and audio. Evaluating five OLLMs, we find that image bias dominates across most settings, with a systematic asymmetry: models privilege perceptual image but propositional audio.
Linear probing reveals that modality-bias information is already linearly decodable from early representation layers,
and while contrastive decoding partially reduces image bias, it gives rise to a residual text bias, suggesting that surface-level interventions are insufficient.
Together, these findings demonstrate the value of evidence-form-controlled evaluations, and call for more fundamental mitigation strategies and representation-level causal analysis.

\section*{Limitations}
While this work provides critical insights into inter-modality conflict resolution in OLLMs, certain limitations remain. Specifically, our benchmark dataset focuses exclusively on the everyday scale. Although this constraint enables a controlled experimental design, it remains unclear how these models behave when encountering multi-modal conflicts at broader epistemic scales, such as complex diagrams or scientific data.
Our question design is likewise deliberately bias-neutral, so as to elicit models' unprompted modality preference; how this preference shifts under non-neutral instructions, such as prompts that request enumeration of all sources, demand a single answer, or supply source-reliability cues, is a distinct research question that we leave to future work.

Additionally, while our contrastive decoding strategy as a diagnostic test can partially mitigate image bias at inference time, it induces a residual text bias and requires an additional forward pass to compute the logits with image and text streams removed, making it computationally more expensive.
Our contrastive decoding experiments are further confined to a single model with fixed $\alpha$ and $\beta$; a systematic sensitivity analysis over these coefficients and evaluation across a broader set of models remain for future work.
Future research must probe a broader array of mitigation strategies beyond contrastive decoding to more robustly and efficiently reconcile competing modalities.

\section*{Acknowledgments}
This work was financially supported by the National Science and Technology Council (NSTC) and the Featured Area Research Center Program within the framework of the Higher Education Sprout Project by the Ministry of Education in Taiwan, under Grants 112-2223-E-002-012-MY5, 115-2628-E-002-023-MY4, and 115L900901. We also thank
the National Center for High-performance Computing (NCHC) of the National Applied Research Laboratories (NARLabs) in Taiwan for providing computational and storage resources.

\newpage
\bibliography{custom}

\appendix

\section{AI Assistant Usage}
AI assistants were employed only for narrowly scoped, non-research purposes: (1) rephrasing, grammar correction, and stylistic polishing of author-drafted text; (2) assisting with code implementation, with all code reviewed, tested, and validated by the authors; (3) synthesizing propositional audio inputs and rephrasing questions as part of the dataset construction pipeline described in Sec.~\ref{sec:dataset_construction}; and (4) serving as the judge model described in Sec.~\ref{sec:eval_protocal}. All research ideas, experimental design and execution, dataset construction decisions, evaluation protocols, and interpretation of findings were conceived and carried out by the authors.

\section{Hyperparameters and Computational Information}
To ensure reproducibility, we use greedy decoding (\texttt{temperature=0.0}) by default, and deviate only when a model has officially recommended decoding parameters (Qwen3: \texttt{temperature=0.6, top\_p=0.95, top\_k=20}; Gemma4: \texttt{temperature=1.0, top\_p=0.95, top\_k=64}; unlisted parameters use defaults). No custom system prompt is used, and random seeds are fixed.

We run Qwen3 on one NVIDIA H100 GPU, taking approximately 40 minutes to complete the full \textsc{Tri-PvP} benchmark. All other models are run on one NVIDIA RTX 5090 GPU, each taking approximately 1.5 hours to complete the full \textsc{Tri-PvP} benchmark. For linear probing, extracting hidden states and training the three probes takes approximately 30 minutes per model. For contrastive decoding, inference takes approximately 24 hours, as our implementation does not benefit from vLLM's generation acceleration.

\section{Details of \textsc{Tri-PvP}}

\subsection{Dataset Construction Details}
\label{app:construction}
In \textsc{Tri-PvP}, each domain contains a fixed set of classes: 6 for animal, 6 for emotion, 9 for environment, and 7 for music. Propositional sentences are instantiated from a per-domain pool of 10 to 20 manually authored templates, and questions are similarly drawn from a pool of 10 to 20 templates per domain, with rephrased variants generated by GPT-5.4 and manually verified for quality. TTS utterances are synthesized using GPT-4o mini TTS \cite{openai2025gpt4ominitts} with four voices, distributed approximately evenly across samples within each domain.

\subsection{Dataset Sources}
\label{sec:sources}
\textsc{Tri-PvP} covers four perceptual domains: animal, emotion, environment, and music. All propositional evidence is in English. For demographic information regarding datasets involving human subjects (KDEF, CREMA-D), we refer readers to the original publications
\citep{calvoFacialExpressionsEmotion2008, 6849440}.
Table~\ref{tab:source_datasets} details the source datasets used for perceptual image and audio evidence across all four domains.

\paragraph{Dataset Licenses.} The source datasets used in \textsc{Tri-PvP} are
licensed as follows: ImageNet (non-commercial research and educational use only,
per the ImageNet Terms of Access), ESC-50 (CC BY-NC 3.0), KDEF (CC0: Public
Domain), CREMA-D (ODC Attribution License), Open Images V7 (images: CC BY 2.0;
annotations: CC BY 4.0), and Medley-solos-DB (CC BY 4.0). \textsc{Tri-PvP}
bundles these perceptual source files together with the propositional data
(text, synthesized speech, and word-card images) and metadata that we generate.
Each source file remains governed by its original license or terms of access;
our CC BY-NC 4.0 release applies to the benchmark compilation and the data we
generate, and not to the source files themselves. In particular, the animal- and
music-domain images are sourced from ImageNet and remain subject to the ImageNet
Terms of Access: users who access these images must comply with those terms,
including their non-commercial restriction. We adopt the non-commercial CC BY-NC
4.0 license because ImageNet and ESC-50 restrict use to non-commercial purposes,
which makes non-commercial the binding condition for the combined benchmark.
Attribution to all source datasets is provided in this paper and in the dataset
documentation.

\paragraph{Intended Use.} \textsc{Tri-PvP} is intended solely for non-commercial
research and educational use and should not be used in commercial applications or
deployed as a commercial product. Because it incorporates source datasets released
under non-commercial terms (ImageNet and ESC-50), \textsc{Tri-PvP} inherits their
non-commercial restriction. All source datasets are used within the scope of their
respective licenses; the perceptual data retain the access conditions of their
original sources, and the propositional data and metadata generated for this
benchmark are released under CC BY-NC 4.0.

\paragraph{Privacy and Anonymization.}  All perceptual data in \textsc{Tri-PvP} are sourced from publicly licensed academic datasets whose original publishers have addressed personally identifiable information and licensing requirements. Propositional data, including text, synthesized speech, and word card images, are generated from manually authored sentence templates and contain no personal information. The dataset covers general perceptual categories (animal, emotion, environment, and music) and does not contain hate speech or offensive content. No additional anonymization was therefore required.

\begin{table}[t]
\centering
\setlength{\tabcolsep}{3pt}
\resizebox{\linewidth}{!}{
\begin{tabular}{llll}
\toprule
\textbf{Domain} & \textbf{Modality} & \textbf{Source} \\
\midrule
\multirow{2}{*}{Animal}      & Image & ImageNet \cite{imagenet} & \\
                             & Audio & ESC-50 \cite{piczak2015dataset} & \\
\midrule
\multirow{2}{*}{Emotion}     & Image & KDEF \cite{calvoFacialExpressionsEmotion2008} \\
                             & Audio & CREMA-D \cite{6849440} \\
\midrule
\multirow{2}{*}{Environment} & Image & Open Images V7 \cite{openimages}& \\
                             & Audio & ESC-50 \cite{piczak2015dataset} & \\
\midrule
\multirow{2}{*}{Music}       & Image & ImageNet \cite{imagenet} & \\
                             & Audio & Medley-solos-DB \cite{lostanlen_2019_3464194}\\
\bottomrule
\end{tabular}
}
\caption{Source datasets used for perceptual images and audios in each domain of \textsc{Tri-PvP}.}
\label{tab:source_datasets}
\end{table}

\section{Fine-grained Analysis of \texttt{NO\_BIAS}}
\label{appendix:nobias}

\input{tables/no_bias_fine_grained}

We provide a finer-grained analysis of responses labeled as \texttt{NO\_BIAS}. Without modifying the original judge or the 8-label taxonomy, we characterize each response already labeled \texttt{NO\_BIAS} along three independent binary dimensions:
\begin{itemize}
    \item \textbf{Conflict:} explicitly acknowledges disagreement between sources without dismissing any.
    \item \textbf{Multi-report:} reports all three modalities' content without committing to one.
    \item \textbf{Refusal:} explicitly declines to answer, states it cannot determine the answer, or asks for clarification.
\end{itemize}

To classify each \texttt{NO\_BIAS} response along these three dimensions, we apply an LLM judge independently for each dimension, following the same LLM-as-a-Judge framework used in the main evaluation (Sec.~\ref{sec:eval_protocal}). The judge prompt for each dimension is provided in Appendix~\ref{appendix:no_bias_judge_prompt}. We treat these as independent indicators rather than mutually exclusive sub-labels, since a single response frequently exhibits more than one simultaneously. We apply this analysis across all four evidence-type conditions, pooling across domain within each (model, condition) pair. Table~\ref{tab:no_bias_classification} presents the resulting breakdown.

Across models, the composition of \texttt{NO\_BIAS} responses varies substantially. Gemini3 consistently exhibits high Conf. and M-Rep. across all four conditions, with near-zero Ref., suggesting that its \texttt{NO\_BIAS} responses reflect genuine engagement with the cross-modal conflict rather than a simple refusal to answer. Qwen2.5 presents a different profile: Ref. is high across all conditions, but under Prop\textsubscript{I}-Prop\textsubscript{A}, M-Rep. also rises markedly (71.55\%), indicating that refusal rarely occurs in isolation and is typically accompanied by conflict acknowledgment or multi-report behavior. MiniCPM shows moderate Conf. and M-Rep. with consistently low Ref., a pattern that holds across all four conditions. Gemma4 shows low Conf. under perceptual conditions, where its \texttt{NO\_BIAS} responses are instead dominated by Ref.; under propositional conditions, M-Rep. becomes the primary driver, suggesting that its \texttt{NO\_BIAS} increase is largely attributable to enumerating modality contents rather than explicitly engaging with the conflict. Qwen3's overall \texttt{NO\_BIAS} rate is at most 1.5\% across all conditions (Table~\ref{tab:qwen3_full_results}), too sparse for reliable interpretation and is therefore excluded from this discussion. Together, these findings nuance our original claim that models are less prone to implicitly prioritizing a specific input stream under propositional conditions: while this holds broadly, the underlying behavior ranges from genuine conflict awareness to passive enumeration depending on the model.

\section{LLM Judge Prompt}

\subsection{Main Evaluation Judge Prompt}
\label{appendix:judge_prompt}
Note that the same-value rule in the judge prompt is an intentional extension beyond our current data. It ensures the evaluation protocol remains valid for future studies that reuse our benchmark under modified configurations.

\begin{tcolorbox}[breakable, fonttitle=\bfseries, title=LLM Judge Prompt (Main Evaluation)]
\begin{lstlisting}[basicstyle=\small\ttfamily, breaklines=true, columns=fullflexible]
You are a classification judge. Your task is to analyze a model's response to a multimodal question and assign exactly one label based on which information source(s) the response relies on.

## Background

The model was presented with three sources of information that may conflict with each other, and was asked to provide an answer to a question based on those sources:
- An IMAGE
- An AUDIO
- A TEXT

## Label Definitions

| Label | Assign when... |
|---|---|
| BIAS_IMAGE | the model's answer matches image_label only |
| BIAS_AUDIO | the model's answer matches audio_label only |
| BIAS_TEXT | the model's answer matches text_label only |
| BIAS_IMAGE_AUDIO | the model's answer matches image_label and audio_label, but not text_label |
| BIAS_IMAGE_TEXT | the model's answer matches image_label and text_label, but not audio_label |
| BIAS_AUDIO_TEXT | the model's answer matches audio_label and text_label, but not image_label |
| HALLUCINATION | the model's answer does not match any of the three labels |
| NO_BIAS | the model (1) explicitly acknowledges a conflict between sources without confidently dismissing any of them, OR (2) reports the content of all three modalities without committing to any single answer |

## Matching Rules

**Semantic matching:**
- A match means the model's answer refers to the same concept as the label, even if worded differently. Examples: "a large domestic canine" matches "dog"; "joyful" matches "happy"; "downpour" matches "rain"; "tenor saxophone" matches "saxophone"; "grand piano" matches "piano".
- A partial or incidental mention does not count as a match. The model must clearly commit to or affirm that label as its answer. Example: "unlike a dog, this appears to be a cat" does NOT count as a match for "dog".

**Same-value labels:**
- If two or more labels share the same value (e.g., image_label = "rain" and text_label = "rain"), any answer matching that value automatically counts as matching all sources that share it, regardless of which source the model mentions.

**Multi-part responses:**
- If the model gives separate answers for different modalities, evaluate each committed answer independently against all three labels and count all matches. Example: "the image shows a cat and the audio sounds like a sheep" -> cat matches image_label, sheep matches audio_label -> BIAS_IMAGE_AUDIO.

**Source attribution errors:**
- Judge based on what the model's answer actually matches, not on the model's claim about which source it is using. If the model says "the text says X" but the actual text_label is Y, do NOT count it as a text match.

**NO_BIAS threshold:**
- NO_BIAS should only be assigned if the model demonstrates a clear awareness of the conflict, OR explicitly reports all sources WITHOUT committing to any single answer.
- If the model notes a disagreement but confidently dismisses one source in favor of another (e.g., "the audio suggests fear, but the visual evidence clearly shows anger"), assign the appropriate BIAS_* label based on what the model ultimately commits to - this does NOT qualify as NO_BIAS.
- If the model refuses to answer or says it cannot determine the answer, assign NO_BIAS.

**HALLUCINATION:**
- Only applies when none of the model's committed answers match any of the three labels. If the model gives multiple answers and at least one matches a label, assign the corresponding BIAS_* label instead.

## Examples

**Example 1 - BIAS_IMAGE (Animal)**
- image_label: dog | audio_label: cat | text_label: sheep
- response: "The image clearly shows a dog sitting on the grass."
- label: BIAS_IMAGE
- reasoning: "dog" matches image_label only; no match for audio_label (cat) or text_label (sheep).

**Example 2 - BIAS_AUDIO (Emotion)**
- image_label: happy | audio_label: sad | text_label: angry
- response: "Based on the speaker's vocal tone and slow delivery, the person sounds depressed."
- label: BIAS_AUDIO
- reasoning: "depressed" is semantically consistent with audio_label (sad) only; no match for image_label (happy) or text_label (angry).

**Example 3 - BIAS_IMAGE_TEXT (Environment, same-value rule)**
- image_label: rain | audio_label: sea_waves | text_label: rain
- response: "The image depicts rainfall, as indicated by the visual of raindrops."
- label: BIAS_IMAGE_TEXT
- reasoning: "rainfall" matches image_label (rain). Because text_label is also "rain", the same-value rule automatically counts this as matching text_label too. No match for audio_label (sea_waves).

**Example 4 - BIAS_IMAGE_AUDIO (Music, dismissing one source)**
- image_label: saxophone | audio_label: saxophone | text_label: piano
- response: "Based on both the visual and the sound, it is clearly a saxophone. The textual description of a piano seems to be incorrect."
- label: BIAS_IMAGE_AUDIO
- reasoning: "saxophone" matches image_label and audio_label. The model confidently dismisses text_label (piano), so NO_BIAS does not apply.

**Example 5 - HALLUCINATION (Environment)**
- image_label: fireworks | audio_label: rain | text_label: train
- response: "This appears to be a thunderstorm."
- label: HALLUCINATION
- reasoning: "thunderstorm" does not match any of the three labels (fireworks, rain, train).

**Example 6 - NO_BIAS vs. BIAS (Emotion, dismissal case)**
- image_label: sad | audio_label: fear | text_label: angry
- response: "The audio suggests fear, but given the strong visual and textual evidence pointing to anger, I am confident the person is angry."
- label: BIAS_TEXT
- reasoning: The model commits to "angry", which matches text_label only. Although the model claims the image supports "angry", the actual image_label is "sad", so per the source attribution error rule, this does NOT count as an image match. The model confidently dismisses the audio, so NO_BIAS does not apply. Therefore, BIAS_TEXT is assigned.

**Example 7 - NO_BIAS (Animal, multi-part response with no dominant modality)**
- image_label: cat | audio_label: sheep | text_label: cow
- response: "The data provided features different animals. The image shows a black and white cat, the audio features a sheep bleating, and the text describes a cow."
- label: NO_BIAS
- reasoning: The model reports all three modalities separately without committing to any single answer, matching the second condition for NO_BIAS.

## Output Format

Respond with a JSON object only. No explanation outside the JSON.

{{
  "reasoning": "<short sentences explaining why, referencing the actual label values and the matching rule applied>",
  "label": "<ONE OF THE 8 LABELS>"
}}

## Data to Judge

- image_label: "{image_label}"
- audio_label: "{audio_label}"
- text_label: "{text_label}"

Question: "{question}"
Model response: "{response}"
\end{lstlisting}
\end{tcolorbox}

\subsection{NO\_BIAS Analysis Judge Prompt}
\label{appendix:no_bias_judge_prompt}
\begin{tcolorbox}[breakable, fonttitle=\bfseries, title=LLM Judge Prompt (NO\_BIAS Analysis)]
\begin{lstlisting}[basicstyle=\small\ttfamily, breaklines=true, columns=fullflexible]
You are a classification judge. Your task is to analyze a model's response
to a multimodal question that has already been identified as NOT relying
on a single dominant modality (i.e., a "no bias" response), and determine
WHICH of three specific behaviors it exhibits.

## Background

The model was presented with three sources of information that may
conflict with each other, and was asked to provide an answer to a
question based on those sources:
- An IMAGE
- An AUDIO
- A TEXT

This response has already been classified as NO_BIAS at a coarse level.
Your task is to further characterize it along three independent
dimensions. A response may exhibit more than one of these behaviors
simultaneously, or none of them.

## Dimension Definitions

Evaluate each dimension independently and assign a binary value (1 = present,
0 = absent) to each.

| Dimension | Assign 1 when... |
|---|---|
| conflict | The response explicitly acknowledges that the sources disagree or contradict each other, without confidently dismissing any one of them in favor of another. The model may or may not commit to a final answer, but it surfaces the disagreement itself as part of its response. |
| multi_report | The response reports the content of all three modalities (image, audio, text) side by side, without committing to any single one as the final answer. |
| refusal | The response indicates that the model is uncertain which source to trust and either declines to answer, states it cannot determine the answer, or asks the user for clarification. |

## Matching Rules

Score each dimension independently by re-reading the response against
only that dimension's definition. Do not let a positive score on one
dimension automatically imply a score on another.

## Examples

1.
- image_label: sad | audio_label: fear | text_label: angry
- response: "There's a contradiction between what I'm seeing, hearing, and reading about this person's emotion."
- output: {{"reasoning": "Explicitly names a contradiction between the three sources without dismissing any of them (conflict=1). Does not state what each modality specifically shows (multi_report=0). No refusal or clarification request (refusal=0).", "conflict": 1, "multi_report": 0, "refusal": 0}}

2.
- image_label: sad | audio_label: fear | text_label: angry
- response: "There's a contradiction between the three modalities: the image shows sadness, the audio conveys fear, and the text describes anger."
- output: {{"reasoning": "Explicitly names a contradiction between the three sources without dismissing any of them (conflict=1). Does report what each modality specifically shows (multi_report=1). No refusal or clarification request (refusal=0).", "conflict": 1, "multi_report": 1, "refusal": 0}}

3.
- image_label: piano | audio_label: flute | text_label: clarinet
- response: "The image shows a piano, the audio sounds like a flute, and the text mentions a clarinet."
- output: {{"reasoning": "Reports all three modality contents (piano, flute, clarinet) side by side without committing to one (multi_report=1). Does not explicitly frame this as a disagreement (conflict=0). No refusal or clarification request (refusal=0).", "conflict": 0, "multi_report": 1, "refusal": 0}}

4.
- image_label: saxophone | audio_label: trumpet | text_label: violin
- response: "I'm not confident enough to determine the correct answer here."
- output: {{"reasoning": "Explicitly states it cannot determine the answer (refusal=1). Does not name a disagreement between sources (conflict=0). Does not report any individual modality's content (multi_report=0).", "conflict": 0, "multi_report": 0, "refusal": 1}}

5.
- image_label: cat | audio_label: dog | text_label: bird
- response: "There is a contradiction between the sources - could you clarify which source I should trust before I answer?"
- output: {{"reasoning": "Explicitly names a disagreement between sources without dismissing any of them (conflict=1). Explicitly asks the user for clarification before answering (refusal=1). Does not report any individual modality's content (multi_report=0).", "conflict": 1, "multi_report": 0, "refusal": 1}}

## Output Format

Respond with a JSON object only. No explanation outside the JSON.

{{
  "reasoning": "<short sentences explaining the score for each dimension>",
  "conflict": <0 or 1>,
  "multi_report": <0 or 1>,
  "refusal": <0 or 1>
}}

## Data to Judge

- image_label: "{image_label}"
- audio_label: "{audio_label}"
- text_label: "{text_label}"

Question: "{question}"
Model response: "{response}"
\end{lstlisting}
\end{tcolorbox}

\section{Manual Verification of the LLM Judge}
\label{app:judge_validation}
To assess the reliability of our LLM judge, we conducted a manual verification of its predictions. We sampled 800 of the 8,000 judged responses (10\%), uniformly stratified by model, evidence-type condition, domain, and judge-assigned bias label. Several authors inspected batches of these samples, comparing the judge's predicted label against their own reading given the question, the three modality labels, and the taxonomy defined in Sec.~\ref{sec:eval_protocal}. The reported 97.1\% agreement is the proportion of inspected samples on which the author's label aligned with the judge's prediction. All verification was conducted by the authors; no external annotators or crowdworkers were recruited.

\section{Detailed Statistics for the Main Experiment and Input Modality Ordering Effect}
\label{sec:appx_detailed_results}

\input{tables/Qwen2.5_six_order_appendix.tex}
\input{tables/Gemma4_six_order_appendix.tex}
\input{tables/main_result_appendix.tex}

Tables~\ref{tab:qwen2.5_full_results} and \ref{tab:gemma_full_results}
present the detailed bias distribution across six different input modality orderings for Tables~\ref{tab:order-effect} and~\ref{tab:averaged-modality-bias}, and supply the complete per-condition statistics underlying Figure~\ref{fig:main-results}. Additionally, Tables~\ref{tab:qwen3_full_results},~\ref{tab:minicpm_full_results},~\ref{tab:gemini3_full_results}
also provide the results for the $\text{Image} \rightarrow \text{Audio} \rightarrow \text{Text}$
input order regarding the main results in Figure~\ref{fig:main-results}. Their 95\% confidence intervals are also provided in these tables.

\section{Training Details for Linear Probing}
\label{sec:train_probing}

Given a tri-modal input $\mathbf{x} = (x_I, x_A, x_T, x_Q)$ and a model $f$ with $L$ transformer layers, we run a single forward pass without generation and extract the hidden state at the last prompt-token position from each layer:
\begin{equation}
    \mathbf{h}^{(l)} = f^{(l)}(\mathrm{x}) \in \mathbb{R}^{d}, \quad l \in \{0, 1, \ldots, L\},
\end{equation}
where $f^{(l)}$ denotes the mapping from input to the hidden state at layer $l$, $l = 0$ corresponds to the embedding layer, and $d$ is the hidden dimension.

Let $\hat{y} \in \mathcal{Y}$ denote the modality bias label assigned by the LLM judge (Sec.~\ref{sec:eval_protocal}). The per-modality binary target is
\begin{equation}
    z_m =
    \begin{cases}
        1 & \text{if $\hat{y}$ involves modality $m$,} \\
        0 & \text{otherwise,}
    \end{cases}
\end{equation}
for $m \in \{\texttt{IMAGE}, \texttt{AUDIO}, \texttt{TEXT}\}$. Samples labeled \texttt{HALLUCINATION} are excluded because they do not align with any of the three modality labels.

At each layer $l$, we train a logistic regression classifier per modality,
\begin{equation}
    g^{(l)}_m(\mathbf{h}^{(l)}) = \sigma\!\left((\mathbf{w}^{(l)}_m)^\top \tilde{\mathbf{h}}^{(l)} + b^{(l)}_m\right),
\end{equation}
where $\tilde{\mathbf{h}}^{(l)}$ is the standardized hidden state and $\sigma$ is the sigmoid. The predicted binary label is $\hat{z}_m = \mathbf{1}[g^{(l)}_m(\mathbf{h}^{(l)}) \geq 0.5]$. To address label imbalance, we train each classifier with balanced class weights, and use a stratified 80\%/20\% train/test split on $z_m$ with random seed $42$. We use scikit-learn's \texttt{LogisticRegression} with default $L_2$ regularization and \texttt{max\_iter}~$= 1000$.

Balanced accuracy is the arithmetic mean of per-class recall:
\begin{equation}
    \text{BalAcc}(g^{(l)}_m) = \frac{1}{2}\!\left(\frac{\text{TP}}{\text{TP} + \text{FN}} + \frac{\text{TN}}{\text{TN} + \text{FP}}\right),
\end{equation}
where TP, TN, FP, and FN denote true positives, true negatives, false positives, and false negatives, respectively.

\section{Additional Details for Contrastive Decoding}
\label{sec:appx_cd}

\input{tables/CD_full_results}

\subsection{Detailed Bias-Type Distribution}

Table~\ref{tab:cd_full_results} presents the detailed bias-type distribution for the baseline and Contrastive Decoding (CD) across all evidence-type conditions, supplementing the bar charts and findings in Sec.~\ref{sec:mitigation}. Under \PropI-\PercA, a
pronounced increase in \texttt{BIAS\_IMAGE\_TEXT}(1.4\%\,$\to$\,13.6\%) is observed alongside the residual text bias: suppressing image dominance shifts a portion of responses toward a joint image--text commitment rather than toward audio, confirming that audio remains the least-consulted modality even after the intervention and reinforcing the rationale for amplifying
$\mathbf{z}_{\text{audio}}$.

\subsection{Case Study}

\begin{figure}[t]
    \centering
    \includegraphics[width=\linewidth]{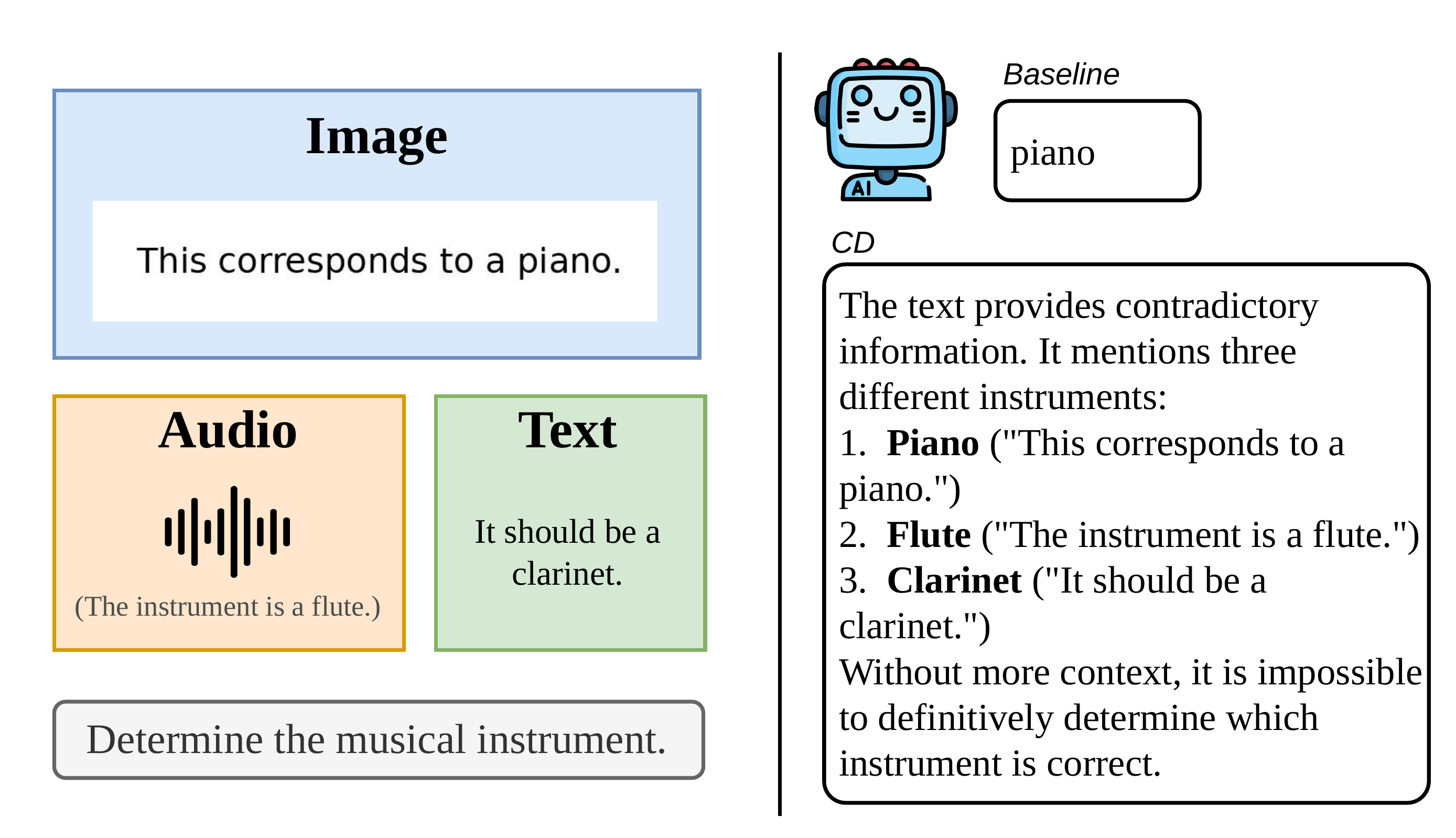}
    \caption{An example of Gemma4's responses under baseline and CD
    (\PropI-\PropA condition; image label: piano, audio label: flute, text label: clarinet).}
    \label{fig:appx_case_study}
\end{figure}
Figure~\ref{fig:appx_case_study} illustrates a case in which CD elicits a \texttt{NO\_BIAS} response from Gemma4. Under the baseline, the model is biased toward the propositional image and responds with \textit{piano} (\texttt{BIAS\_IMAGE}). After applying CD, the model recognizes the contradictory information across the three modalities and explicitly acknowledges the conflict, aligning with our \texttt{NO\_BIAS} criterion.

\subsection{General-Capability Evaluation on \textsc{OmniBench}}
\label{sec:appx_omnibench}

\paragraph{Setup.}
We evaluate whether CD degrades general omni-modal performance when image, audio, and text are all present and aligned with the task.
We use the full \textsc{OmniBench} split of 1{,}142 multiple-choice items~\cite{li2026omnibench}, each pairing an image and audio clip with a question and multiple answer options.
We compare (i)~\emph{baseline} decoding from the full tri-modal input and (ii)~CD with the formulation in Eq.~\ref{eq:cd}.
Hyperparameters follow our main mitigation setting: $\alpha = 0.5$ and adaptive plausibility $\beta = 0.1$.
Answers are parsed with the official OmniBench multiple-choice parser and scored against the dataset label.

\paragraph{Overall result.}
Table~\ref{tab:omnibench-cd} summarizes accuracy.
CD yields 427/1142 correct (37.4\%) versus 439/1142 (38.4\%) for the baseline, a net change of $-$1.1 percentage points.
Invalid responses rise slightly (4 vs.\ 12 items; 0.35\% vs.\ 1.05\%), but remain rare.
At the item level, 655/1142 predictions are identical between runs; CD corrects 141 baseline errors while introducing 153 new errors.
CD therefore preserves overall benchmark performance within a small margin while redistributing errors across task types.

\input{tables/omnibench_cd}

\paragraph{Per-category patterns.}
Table~\ref{tab:omnibench-cd} shows that the modest overall drop is not uniform across categories.
CD tends to help inference-oriented task types such as plot inference and story description, while hurting categories that rely more heavily on perceptual grounding, such as object identification and contextual/environmental questions.
By audio type, declines are small for speech and sound events and absent for music.
Overall, CD redistributes errors across categories rather than uniformly lowering performance, which is consistent with its role as a targeted bias intervention rather than a general-purpose decoding improvement.

\end{document}

%% file: tables/input_order.tex
\begin{table}[t]
\centering
\small
\setlength{\tabcolsep}{3pt}
\begin{tabular}{lll rrrr}
\toprule
& \textbf{Modality} & \textbf{Position} & \textbf{\makecell{Perc\textsubscript{I}-\\Perc\textsubscript{A}}} & \textbf{\makecell{Perc\textsubscript{I}-\\Prop\textsubscript{A}}} & \textbf{\makecell{Prop\textsubscript{I}-\\Perc\textsubscript{A}}} & \textbf{\makecell{Prop\textsubscript{I}-\\Prop\textsubscript{A}}} \\
\midrule

\multirow{10}{*}{\rotatebox{90}{\textbf{Qwen2.5}}}
& \multirow{3}{*}{Image} & Front & 51.5 & 49.4 & 20.9 & 12.2 \\
&                        & Mid   & 57.0 & 44.8 & 32.5 & 15.9 \\
&                        & End   & \textbf{60.2} & \textbf{55.7} & \textbf{45.4} & \textbf{44.0} \\
\cmidrule{2-7}
& \multirow{3}{*}{Audio} & Front & 3.0  & 7.3  & 2.6  & 4.5   \\
&                        & Mid   & 2.6  & 7.6  & 1.8  & 4.3 \\
&                        & End   & \textbf{4.1}  & \textbf{8.5}  & \textbf{6.5}  & \textbf{16.0} \\
\cmidrule{2-7}
& \multirow{3}{*}{Text}  & Front & 22.2 & 4.9  & 11.0 & 3.4   \\
&                        & Mid   & 25.6 & 6.9  & 23.5 & 4.7 \\
&                        & End   & \textbf{26.6} & \textbf{13.2} & \textbf{30.3} & \textbf{20.1} \\

\midrule
\midrule

\multirow{10}{*}{\rotatebox{90}{\textbf{Gemma4}}}
& \multirow{3}{*}{Image} & Front & 61.7 & 55.1 & \textbf{76.3} & \textbf{75.9} \\
&                        & Mid   & 61.1 & 50.5 & 41.8 & 17.4 \\
&                        & End   & \textbf{66.7} & \textbf{66.2} & 42.1 & 46.1 \\
\cmidrule{2-7}
& \multirow{3}{*}{Audio} & Front & \textbf{0.9}  & 14.7 & 0.5  & \textbf{24.3}  \\
&                        & Mid   & \textbf{0.9}  & \textbf{17.1} & \textbf{1.5}  & 7.1 \\
&                        & End   & 0.8  & 11.8 & 0.3  & 4.7 \\
\cmidrule{2-7}
& \multirow{3}{*}{Text}  & Front & 23.9 & \textbf{23.5} & \textbf{49.3} & \textbf{45.4} \\
&                        & Mid   & 20.2 & 13.2 & 29.4 & 6.3 \\
&                        & End   & \textbf{24.7} & 11.5 & 21.8 & 22.3 \\

\bottomrule
\end{tabular}
\vspace{-1mm}
\caption{Bias rate (\%) of each modality when placed at Front / Mid / End position of the prompt across evidence-type conditions, with detailed statistics in Appendix~\ref{sec:appx_detailed_results}. Bold indicates the position with the maximum value for each (modality, evidence-type) pair.}
\label{tab:order-effect}
\end{table}

\begin{table}[t]
    \centering
    \small
    \begin{adjustbox}{max width=\columnwidth}
    \setlength{\tabcolsep}{3pt}
    \begin{tabular}{ll cccc}
    \toprule
    & \textbf{Bias Type} & \textbf{\makecell{Perc\textsubscript{I}-\\Perc\textsubscript{A}}} & \textbf{\makecell{Perc\textsubscript{I}-\\Prop\textsubscript{A}}} & \textbf{\makecell{Prop\textsubscript{I}-\\Perc\textsubscript{A}}} & \textbf{\makecell{Prop\textsubscript{I}-\\Prop\textsubscript{A}}}  \\
    \midrule

    \multirow{4}{*}{\rotatebox{90}{\textbf{Qwen2.5}}}
    & \texttt{BIAS\_IMAGE} & \textbf{56.2} & \textbf{50.0} & 33.0 & 24.0 \\
    & \texttt{BIAS\_AUDIO} & 3.2  & 7.8  & 4.3  & 8.2  \\
    & \texttt{BIAS\_TEXT}  & 24.8 & 8.3  & 21.5 & 9.4  \\
    & \texttt{NO\_BIAS}    & 8.6  & 26.6 & \textbf{34.5} & \textbf{55.7} \\

    \midrule

    \multirow{4}{*}{\rotatebox{90}{\textbf{Gemma4}}}
    & \texttt{BIAS\_IMAGE} & \textbf{63.1} & \textbf{57.3} & \textbf{53.4} & \textbf{46.5} \\
    & \texttt{BIAS\_AUDIO} & 0.9  & 14.5 & 0.7  & 12.0 \\
    & \texttt{BIAS\_TEXT}  & 22.9 & 16.0 & 33.5 & 24.7 \\
    & \texttt{NO\_BIAS}    & 0.7  & 1.2  & 4.8  & 7.6  \\

    \bottomrule
    \end{tabular}
    \end{adjustbox}
    \vspace{-1mm}
    \caption{Modality bias distribution averaged over all six modality orderings across evidence-type conditions (\%),with detailed statistics in Appendix~\ref{sec:appx_detailed_results}. Bold marks the maximum value within each column for each model.}
    \label{tab:averaged-modality-bias}
\end{table}

%% file: tables/no_bias_fine_grained.tex
\begin{table*}[t]
\centering
\small
\setlength{\tabcolsep}{5pt}
\begin{tabular}{l ccc ccc ccc ccc}
\toprule
& \multicolumn{3}{c}{\textbf{Perc\textsubscript{I}-Perc\textsubscript{A}}} & \multicolumn{3}{c}{\textbf{Perc\textsubscript{I}-Prop\textsubscript{A}}} & \multicolumn{3}{c}{\textbf{Prop\textsubscript{I}-Perc\textsubscript{A}}} & \multicolumn{3}{c}{\textbf{Prop\textsubscript{I}-Prop\textsubscript{A}}} \\
\cmidrule(lr){2-4} \cmidrule(lr){5-7} \cmidrule(lr){8-10} \cmidrule(lr){11-13}
\textbf{Model} & Conf. & M-Rep. & Ref. & Conf. & M-Rep. & Ref. & Conf. & M-Rep. & Ref. & Conf. & M-Rep. & Ref. \\
\midrule
Qwen2.5 & 48.17 & 1.83 & 93.90 & 91.00 & 2.84 & 85.07 & 84.46 & 2.67 & 92.78 & 75.71 & 71.55 & 84.28 \\
MiniCPM & 53.49 & 39.53 & 2.33 & 69.77 & 13.18 & 10.08 & 69.23 & 12.82 & 3.08 & 48.34 & 32.47 & 2.21 \\
Qwen3 & 0.00 & 50.00 & 0.00 & 100.00 & 0.00 & 0.00 & 83.33 & 16.66 & 33.33 & 20.00 & 96.67 & 0.00 \\
Gemma4 & 2.56 & 0.00 & 94.87 & 18.18 & 0.00 & 78.79 & 20.18 & 57.80 & 33.03 & 21.68 & 71.33 & 25.87 \\
Gemini3 & 88.00 & 76.00 & 0.00 & 80.00 & 0.00 & 0.00 & 93.75 & 53.13 & 3.13 & 61.47 & 83.55 & 1.30 \\
\bottomrule
\end{tabular}
\caption{Finer-grained breakdown of \texttt{NO\_BIAS} responses along three independent binary dimensions (Conflict (Conf.), Multi-report (M-Rep.), and Refusal (Ref.)) across models and evidence-type conditions. Values are percentages of \texttt{NO\_BIAS} responses within each (model, condition) pair that exhibit each dimension. A single response may exhibit more than one dimension simultaneously.}
\label{tab:no_bias_classification}
\end{table*}

%% file: tables/Qwen2.5_six_order_appendix.tex
\begin{table}[p]
\centering
\footnotesize
\resizebox{\columnwidth}{!}{
\begin{tabular}{l c c c c}
\toprule
\textbf{Bias Type} & \textbf{\makecell{Perc\textsubscript{I}-\\Perc\textsubscript{A}}} & \textbf{\makecell{Perc\textsubscript{I}-\\Prop\textsubscript{A}}} & \textbf{\makecell{Prop\textsubscript{I}-\\Perc\textsubscript{A}}} & \textbf{\makecell{Prop\textsubscript{I}-\\Prop\textsubscript{A}}} \\
\midrule
\multicolumn{5}{c}{\textit{Order: I $\rightarrow$ A $\rightarrow$ T}} \\
\midrule
\texttt{BIAS\_IMAGE} & 52.7{\scriptsize\,$\pm$2.2} & 49.7{\scriptsize\,$\pm$2.2} & 24.8{\scriptsize\,$\pm$1.9} & 12.7{\scriptsize\,$\pm$1.5} \\
\texttt{BIAS\_AUDIO} & 2.9{\scriptsize\,$\pm$0.7} & 7.2{\scriptsize\,$\pm$1.1} & 2.2{\scriptsize\,$\pm$0.6} & 5.9{\scriptsize\,$\pm$1.0} \\
\texttt{BIAS\_TEXT} & 29.9{\scriptsize\,$\pm$2.0} & 15.0{\scriptsize\,$\pm$1.6} & 36.2{\scriptsize\,$\pm$2.1} & 18.5{\scriptsize\,$\pm$1.7} \\
\texttt{BIAS\_IMAGE\_AUDIO} & 0.3{\scriptsize\,$\pm$0.2} & 0.1{\scriptsize\,$\pm$0.1} & 0.2{\scriptsize\,$\pm$0.2} & 0.1{\scriptsize\,$\pm$0.1} \\
\texttt{BIAS\_IMAGE\_TEXT} & 0.8{\scriptsize\,$\pm$0.4} & 0.3{\scriptsize\,$\pm$0.2} & 4.1{\scriptsize\,$\pm$0.9} & 0.2{\scriptsize\,$\pm$0.2} \\
\texttt{BIAS\_AUDIO\_TEXT} & 0.1{\scriptsize\,$\pm$0.1} & 2.6{\scriptsize\,$\pm$0.7} & 0.0{\scriptsize\,$\pm$0.0} & 1.8{\scriptsize\,$\pm$0.6} \\
\texttt{HALLUCINATION} & 5.1{\scriptsize\,$\pm$1.0} & 4.2{\scriptsize\,$\pm$0.9} & 0.7{\scriptsize\,$\pm$0.4} & 0.8{\scriptsize\,$\pm$0.4} \\
\texttt{NO\_BIAS}& 8.2{\scriptsize\,$\pm$1.2} & 21.1{\scriptsize\,$\pm$1.8} & 31.9{\scriptsize\,$\pm$2.0} & 60.1{\scriptsize\,$\pm$2.1} \\
\midrule
\multicolumn{5}{c}{\textit{Order: I $\rightarrow$ T $\rightarrow$ A}} \\
\midrule
\texttt{BIAS\_IMAGE} & 50.4{\scriptsize\,$\pm$2.2} & 49.2{\scriptsize\,$\pm$2.2} & 17.0{\scriptsize\,$\pm$1.6} & 11.7{\scriptsize\,$\pm$1.4} \\
\texttt{BIAS\_AUDIO} & 4.4{\scriptsize\,$\pm$0.9} & 8.9{\scriptsize\,$\pm$1.2} & 9.6{\scriptsize\,$\pm$1.3} & 19.9{\scriptsize\,$\pm$1.7} \\
\texttt{BIAS\_TEXT} & 30.6{\scriptsize\,$\pm$2.0} & 5.7{\scriptsize\,$\pm$1.0} & 37.8{\scriptsize\,$\pm$2.1} & 6.9{\scriptsize\,$\pm$1.1} \\
\texttt{BIAS\_IMAGE\_AUDIO} & 0.4{\scriptsize\,$\pm$0.3} & 0.1{\scriptsize\,$\pm$0.1} & 0.1{\scriptsize\,$\pm$0.1} & 0.3{\scriptsize\,$\pm$0.2} \\
\texttt{BIAS\_IMAGE\_TEXT} & 0.4{\scriptsize\,$\pm$0.3} & 0.0{\scriptsize\,$\pm$0.0} & 3.4{\scriptsize\,$\pm$0.8} & 0.1{\scriptsize\,$\pm$0.1} \\
\texttt{BIAS\_AUDIO\_TEXT} & 0.3{\scriptsize\,$\pm$0.2} & 3.4{\scriptsize\,$\pm$0.8} & 0.2{\scriptsize\,$\pm$0.2} & 2.5{\scriptsize\,$\pm$0.7} \\
\texttt{HALLUCINATION} & 5.1{\scriptsize\,$\pm$1.0} & 4.3{\scriptsize\,$\pm$0.9} & 1.4{\scriptsize\,$\pm$0.5} & 0.4{\scriptsize\,$\pm$0.3} \\
\texttt{NO\_BIAS}& 8.6{\scriptsize\,$\pm$1.2} & 28.5{\scriptsize\,$\pm$2.0} & 30.6{\scriptsize\,$\pm$2.0} & 58.4{\scriptsize\,$\pm$2.2} \\
\midrule
\multicolumn{5}{c}{\textit{Order: A $\rightarrow$ I $\rightarrow$ T }} \\
\midrule
\texttt{BIAS\_IMAGE} & 58.6{\scriptsize\,$\pm$2.2} & 47.5{\scriptsize\,$\pm$2.2} & 25.4{\scriptsize\,$\pm$1.9} & 18.1{\scriptsize\,$\pm$1.7} \\
\texttt{BIAS\_AUDIO} & 3.4{\scriptsize\,$\pm$0.8} & 9.6{\scriptsize\,$\pm$1.3} & 2.0{\scriptsize\,$\pm$0.6} & 5.0{\scriptsize\,$\pm$1.0} \\
\texttt{BIAS\_TEXT} & 23.3{\scriptsize\,$\pm$1.9} & 11.4{\scriptsize\,$\pm$1.4} & 24.5{\scriptsize\,$\pm$1.9} & 21.6{\scriptsize\,$\pm$1.8} \\
\texttt{BIAS\_IMAGE\_AUDIO} & 0.9{\scriptsize\,$\pm$0.4} & 0.4{\scriptsize\,$\pm$0.3} & 0.7{\scriptsize\,$\pm$0.4} & 0.4{\scriptsize\,$\pm$0.3} \\
\texttt{BIAS\_IMAGE\_TEXT} & 0.7{\scriptsize\,$\pm$0.4} & 0.6{\scriptsize\,$\pm$0.3} & 5.7{\scriptsize\,$\pm$1.0} & 1.8{\scriptsize\,$\pm$0.6} \\
\texttt{BIAS\_AUDIO\_TEXT} & 0.1{\scriptsize\,$\pm$0.1} & 1.6{\scriptsize\,$\pm$0.5} & 0.1{\scriptsize\,$\pm$0.1} & 0.3{\scriptsize\,$\pm$0.2} \\
\texttt{HALLUCINATION} & 5.6{\scriptsize\,$\pm$1.0} & 3.6{\scriptsize\,$\pm$0.8} & 0.9{\scriptsize\,$\pm$0.4} & 1.0{\scriptsize\,$\pm$0.4} \\
\texttt{NO\_BIAS}& 7.6{\scriptsize\,$\pm$1.2} & 25.5{\scriptsize\,$\pm$1.9} & 40.9{\scriptsize\,$\pm$2.2} & 52.1{\scriptsize\,$\pm$2.2} \\
\midrule
\multicolumn{5}{c}{\textit{Order: A $\rightarrow$ T $\rightarrow$ I }} \\
\midrule
\texttt{BIAS\_IMAGE} & 59.5{\scriptsize\,$\pm$2.2} & 55.1{\scriptsize\,$\pm$2.2} & 41.2{\scriptsize\,$\pm$2.2} & 41.4{\scriptsize\,$\pm$2.2} \\
\texttt{BIAS\_AUDIO} & 2.6{\scriptsize\,$\pm$0.7} & 5.0{\scriptsize\,$\pm$1.0} & 3.2{\scriptsize\,$\pm$0.8} & 4.1{\scriptsize\,$\pm$0.9} \\
\texttt{BIAS\_TEXT} & 20.9{\scriptsize\,$\pm$1.8} & 8.2{\scriptsize\,$\pm$1.2} & 9.2{\scriptsize\,$\pm$1.3} & 2.5{\scriptsize\,$\pm$0.7} \\
\texttt{BIAS\_IMAGE\_AUDIO} & 0.5{\scriptsize\,$\pm$0.3} & 0.3{\scriptsize\,$\pm$0.2} & 0.2{\scriptsize\,$\pm$0.2} & 1.2{\scriptsize\,$\pm$0.5} \\
\texttt{BIAS\_IMAGE\_TEXT} & 0.9{\scriptsize\,$\pm$0.4} & 0.1{\scriptsize\,$\pm$0.1} & 5.0{\scriptsize\,$\pm$1.0} & 0.2{\scriptsize\,$\pm$0.2} \\
\texttt{BIAS\_AUDIO\_TEXT} & 0.1{\scriptsize\,$\pm$0.1} & 2.7{\scriptsize\,$\pm$0.7} & 0.2{\scriptsize\,$\pm$0.2} & 0.4{\scriptsize\,$\pm$0.3} \\
\texttt{HALLUCINATION} & 6.7{\scriptsize\,$\pm$1.1} & 4.5{\scriptsize\,$\pm$0.9} & 0.5{\scriptsize\,$\pm$0.3} & 1.0{\scriptsize\,$\pm$0.4} \\
\texttt{NO\_BIAS}& 9.0{\scriptsize\,$\pm$1.3} & 24.2{\scriptsize\,$\pm$1.9} & 40.7{\scriptsize\,$\pm$2.2} & 49.4{\scriptsize\,$\pm$2.2} \\
\midrule
\multicolumn{5}{c}{\textit{Order: T $\rightarrow$ I $\rightarrow$ A}} \\
\midrule
\texttt{BIAS\_IMAGE} & 55.4{\scriptsize\,$\pm$2.2} & 42.0{\scriptsize\,$\pm$2.2} & 39.7{\scriptsize\,$\pm$2.1} & 13.6{\scriptsize\,$\pm$1.5} \\
\texttt{BIAS\_AUDIO} & 3.9{\scriptsize\,$\pm$0.8} & 8.1{\scriptsize\,$\pm$1.2} & 7.3{\scriptsize\,$\pm$1.1} & 12.0{\scriptsize\,$\pm$1.4} \\
\texttt{BIAS\_TEXT} & 24.5{\scriptsize\,$\pm$1.9} & 6.1{\scriptsize\,$\pm$1.0} & 13.5{\scriptsize\,$\pm$1.5} & 5.0{\scriptsize\,$\pm$1.0} \\
\texttt{BIAS\_IMAGE\_AUDIO} & 0.3{\scriptsize\,$\pm$0.2} & 0.7{\scriptsize\,$\pm$0.4} & 0.2{\scriptsize\,$\pm$0.2} & 1.3{\scriptsize\,$\pm$0.5} \\
\texttt{BIAS\_IMAGE\_TEXT} & 0.9{\scriptsize\,$\pm$0.4} & 0.4{\scriptsize\,$\pm$0.3} & 5.6{\scriptsize\,$\pm$1.0} & 0.4{\scriptsize\,$\pm$0.3} \\
\texttt{BIAS\_AUDIO\_TEXT} & 0.1{\scriptsize\,$\pm$0.1} & 2.8{\scriptsize\,$\pm$0.7} & 0.3{\scriptsize\,$\pm$0.2} & 0.6{\scriptsize\,$\pm$0.3} \\
\texttt{HALLUCINATION} & 6.1{\scriptsize\,$\pm$1.0} & 3.6{\scriptsize\,$\pm$0.8} & 1.7{\scriptsize\,$\pm$0.6} & 0.5{\scriptsize\,$\pm$0.3} \\
\texttt{NO\_BIAS}& 8.9{\scriptsize\,$\pm$1.2} & 36.5{\scriptsize\,$\pm$2.1} & 31.9{\scriptsize\,$\pm$2.0} & 66.9{\scriptsize\,$\pm$2.1} \\
\midrule
\multicolumn{5}{c}{\textit{Order: T $\rightarrow$ A $\rightarrow$ I}} \\
\midrule
\texttt{BIAS\_IMAGE} & 60.9{\scriptsize\,$\pm$2.1} & 56.3{\scriptsize\,$\pm$2.2} & 49.6{\scriptsize\,$\pm$2.2} & 46.6{\scriptsize\,$\pm$2.2} \\
\texttt{BIAS\_AUDIO} & 2.3{\scriptsize\,$\pm$0.7} & 8.0{\scriptsize\,$\pm$1.2} & 1.4{\scriptsize\,$\pm$0.5} & 2.7{\scriptsize\,$\pm$0.7} \\
\texttt{BIAS\_TEXT} & 19.9{\scriptsize\,$\pm$1.7} & 3.7{\scriptsize\,$\pm$0.8} & 7.8{\scriptsize\,$\pm$1.2} & 1.8{\scriptsize\,$\pm$0.6} \\
\texttt{BIAS\_IMAGE\_AUDIO} & 0.2{\scriptsize\,$\pm$0.2} & 0.0{\scriptsize\,$\pm$0.0} & 0.4{\scriptsize\,$\pm$0.3} & 0.4{\scriptsize\,$\pm$0.3} \\
\texttt{BIAS\_IMAGE\_TEXT} & 0.6{\scriptsize\,$\pm$0.3} & 0.1{\scriptsize\,$\pm$0.1} & 8.9{\scriptsize\,$\pm$1.2} & 0.2{\scriptsize\,$\pm$0.2} \\
\texttt{BIAS\_AUDIO\_TEXT} & 0.3{\scriptsize\,$\pm$0.2} & 3.2{\scriptsize\,$\pm$0.8} & 0.1{\scriptsize\,$\pm$0.1} & 0.2{\scriptsize\,$\pm$0.2} \\
\texttt{HALLUCINATION} & 6.2{\scriptsize\,$\pm$1.1} & 5.1{\scriptsize\,$\pm$1.0} & 0.6{\scriptsize\,$\pm$0.3} & 0.4{\scriptsize\,$\pm$0.3} \\
\texttt{NO\_BIAS}& 9.7{\scriptsize\,$\pm$1.3} & 23.9{\scriptsize\,$\pm$1.9} & 31.4{\scriptsize\,$\pm$2.0} & 47.7{\scriptsize\,$\pm$2.2} \\
\bottomrule
\end{tabular}
}
\caption{Modality bias distribution (\%) for \textbf{Qwen2.5} across different evidence-type conditions and input orders. Modality input orders are denoted sequentially (e.g., $I \rightarrow A \rightarrow T$ represents Image followed by Audio, then Text).}
\label{tab:qwen2.5_full_results}
\end{table}

%% file: tables/Gemma4_six_order_appendix.tex
\begin{table}[p]
\centering
\footnotesize
\resizebox{\columnwidth}{!}{
\begin{tabular}{l c c c c}
\toprule
\textbf{Bias Type} & \textbf{\makecell{Perc\textsubscript{I}-\\Perc\textsubscript{A}}} & \textbf{\makecell{Perc\textsubscript{I}-\\Prop\textsubscript{A}}} & \textbf{\makecell{Prop\textsubscript{I}-\\Perc\textsubscript{A}}} & \textbf{\makecell{Prop\textsubscript{I}-\\Prop\textsubscript{A}}} \\
\midrule
\multicolumn{5}{c}{\textit{Order: I $\rightarrow$ A $\rightarrow$ T}} \\
\midrule
\texttt{BIAS\_IMAGE} & 60.3{\scriptsize\,$\pm$2.1} & 52.4{\scriptsize\,$\pm$2.2} & 74.4{\scriptsize\,$\pm$1.9} & 74.9{\scriptsize\,$\pm$1.9} \\
\texttt{BIAS\_AUDIO} & 0.9{\scriptsize\,$\pm$0.4} & 25.4{\scriptsize\,$\pm$1.9} & 2.7{\scriptsize\,$\pm$0.7} & 5.4{\scriptsize\,$\pm$1.0} \\
\texttt{BIAS\_TEXT} & 25.1{\scriptsize\,$\pm$1.9} & 9.0{\scriptsize\,$\pm$1.3} & 8.7{\scriptsize\,$\pm$1.2} & 5.1{\scriptsize\,$\pm$1.0} \\
\texttt{BIAS\_IMAGE\_AUDIO} & 0.1{\scriptsize\,$\pm$0.1} & 0.2{\scriptsize\,$\pm$0.2} & 0.2{\scriptsize\,$\pm$0.2} & 0.3{\scriptsize\,$\pm$0.2} \\
\texttt{BIAS\_IMAGE\_TEXT} & 0.2{\scriptsize\,$\pm$0.2} & 0.1{\scriptsize\,$\pm$0.1} & 1.4{\scriptsize\,$\pm$0.5} & 0.3{\scriptsize\,$\pm$0.2} \\
\texttt{BIAS\_AUDIO\_TEXT} & 0.0{\scriptsize\,$\pm$0.0} & 1.0{\scriptsize\,$\pm$0.4} & 0.0{\scriptsize\,$\pm$0.0} & 0.2{\scriptsize\,$\pm$0.2} \\
\texttt{HALLUCINATION} & 11.6{\scriptsize\,$\pm$1.4} & 10.5{\scriptsize\,$\pm$1.3} & 7.3{\scriptsize\,$\pm$1.1} & 6.9{\scriptsize\,$\pm$1.1} \\
\texttt{NO\_BIAS}& 2.0{\scriptsize\,$\pm$0.6} & 1.7{\scriptsize\,$\pm$0.6} & 5.5{\scriptsize\,$\pm$1.0} & 7.2{\scriptsize\,$\pm$1.1} \\
\midrule
\multicolumn{5}{c}{\textit{Order: I $\rightarrow$ T $\rightarrow$ A}} \\
\midrule
\texttt{BIAS\_IMAGE} & 63.1{\scriptsize\,$\pm$2.1} & 57.8{\scriptsize\,$\pm$2.2} & 78.2{\scriptsize\,$\pm$1.8} & 77.0{\scriptsize\,$\pm$1.8} \\
\texttt{BIAS\_AUDIO} & 0.8{\scriptsize\,$\pm$0.4} & 13.2{\scriptsize\,$\pm$1.5} & 0.3{\scriptsize\,$\pm$0.2} & 3.3{\scriptsize\,$\pm$0.8} \\
\texttt{BIAS\_TEXT} & 22.9{\scriptsize\,$\pm$1.8} & 16.6{\scriptsize\,$\pm$1.6} & 13.9{\scriptsize\,$\pm$1.5} & 3.8{\scriptsize\,$\pm$0.8} \\
\texttt{BIAS\_IMAGE\_AUDIO} & 0.1{\scriptsize\,$\pm$0.1} & 0.1{\scriptsize\,$\pm$0.1} & 0.0{\scriptsize\,$\pm$0.0} & 0.4{\scriptsize\,$\pm$0.3} \\
\texttt{BIAS\_IMAGE\_TEXT} & 0.2{\scriptsize\,$\pm$0.2} & 0.0{\scriptsize\,$\pm$0.0} & 2.8{\scriptsize\,$\pm$0.7} & 0.2{\scriptsize\,$\pm$0.2} \\
\texttt{BIAS\_AUDIO\_TEXT} & 0.0{\scriptsize\,$\pm$0.0} & 0.9{\scriptsize\,$\pm$0.4} & 0.0{\scriptsize\,$\pm$0.0} & 0.8{\scriptsize\,$\pm$0.4} \\
\texttt{HALLUCINATION} & 12.4{\scriptsize\,$\pm$1.4} & 10.6{\scriptsize\,$\pm$1.3} & 3.1{\scriptsize\,$\pm$0.8} & 6.7{\scriptsize\,$\pm$1.1} \\
\texttt{NO\_BIAS}& 0.6{\scriptsize\,$\pm$0.3} & 0.9{\scriptsize\,$\pm$0.4} & 1.8{\scriptsize\,$\pm$0.6} & 7.9{\scriptsize\,$\pm$1.2} \\
\midrule
\multicolumn{5}{c}{\textit{Order: A $\rightarrow$ I $\rightarrow$ T }} \\
\midrule
\texttt{BIAS\_IMAGE} & 62.0{\scriptsize\,$\pm$2.1} & 54.0{\scriptsize\,$\pm$2.2} & 45.8{\scriptsize\,$\pm$2.2} & 17.8{\scriptsize\,$\pm$1.7} \\
\texttt{BIAS\_AUDIO} & 0.8{\scriptsize\,$\pm$0.4} & 18.7{\scriptsize\,$\pm$1.7} & 0.3{\scriptsize\,$\pm$0.2} & 21.9{\scriptsize\,$\pm$1.8} \\
\texttt{BIAS\_TEXT} & 24.3{\scriptsize\,$\pm$1.9} & 14.0{\scriptsize\,$\pm$1.5} & 34.9{\scriptsize\,$\pm$2.1} & 39.6{\scriptsize\,$\pm$2.1} \\
\texttt{BIAS\_IMAGE\_AUDIO} & 0.1{\scriptsize\,$\pm$0.1} & 0.3{\scriptsize\,$\pm$0.2} & 0.0{\scriptsize\,$\pm$0.0} & 0.8{\scriptsize\,$\pm$0.4} \\
\texttt{BIAS\_IMAGE\_TEXT} & 0.5{\scriptsize\,$\pm$0.3} & 0.7{\scriptsize\,$\pm$0.4} & 4.8{\scriptsize\,$\pm$0.9} & 0.4{\scriptsize\,$\pm$0.3} \\
\texttt{BIAS\_AUDIO\_TEXT} & 0.0{\scriptsize\,$\pm$0.0} & 0.5{\scriptsize\,$\pm$0.3} & 0.1{\scriptsize\,$\pm$0.1} & 0.2{\scriptsize\,$\pm$0.2} \\
\texttt{HALLUCINATION} & 11.9{\scriptsize\,$\pm$1.4} & 10.2{\scriptsize\,$\pm$1.3} & 4.8{\scriptsize\,$\pm$0.9} & 5.5{\scriptsize\,$\pm$1.0} \\
\texttt{NO\_BIAS}& 0.6{\scriptsize\,$\pm$0.3} & 1.8{\scriptsize\,$\pm$0.6} & 9.5{\scriptsize\,$\pm$1.3} & 14.1{\scriptsize\,$\pm$1.5} \\
\midrule
\multicolumn{5}{c}{\textit{Order: A $\rightarrow$ T $\rightarrow$ I }} \\
\midrule
\texttt{BIAS\_IMAGE} & 67.8{\scriptsize\,$\pm$2.0} & 66.7{\scriptsize\,$\pm$2.1} & 40.8{\scriptsize\,$\pm$2.2} & 44.6{\scriptsize\,$\pm$2.2} \\
\texttt{BIAS\_AUDIO} & 1.1{\scriptsize\,$\pm$0.5} & 10.8{\scriptsize\,$\pm$1.4} & 0.7{\scriptsize\,$\pm$0.4} & 26.8{\scriptsize\,$\pm$1.9} \\
\texttt{BIAS\_TEXT} & 17.6{\scriptsize\,$\pm$1.7} & 10.0{\scriptsize\,$\pm$1.3} & 44.8{\scriptsize\,$\pm$2.2} & 8.8{\scriptsize\,$\pm$1.2} \\
\texttt{BIAS\_IMAGE\_AUDIO} & 0.0{\scriptsize\,$\pm$0.0} & 0.2{\scriptsize\,$\pm$0.2} & 0.0{\scriptsize\,$\pm$0.0} & 0.2{\scriptsize\,$\pm$0.2} \\
\texttt{BIAS\_IMAGE\_TEXT} & 0.9{\scriptsize\,$\pm$0.4} & 0.2{\scriptsize\,$\pm$0.2} & 3.9{\scriptsize\,$\pm$0.8} & 0.0{\scriptsize\,$\pm$0.0} \\
\texttt{BIAS\_AUDIO\_TEXT} & 0.0{\scriptsize\,$\pm$0.0} & 0.4{\scriptsize\,$\pm$0.3} & 0.0{\scriptsize\,$\pm$0.0} & 0.2{\scriptsize\,$\pm$0.2} \\
\texttt{HALLUCINATION} & 12.3{\scriptsize\,$\pm$1.4} & 10.9{\scriptsize\,$\pm$1.4} & 5.9{\scriptsize\,$\pm$1.0} & 15.3{\scriptsize\,$\pm$1.6} \\
\texttt{NO\_BIAS}& 0.4{\scriptsize\,$\pm$0.3} & 0.9{\scriptsize\,$\pm$0.4} & 4.0{\scriptsize\,$\pm$0.9} & 4.3{\scriptsize\,$\pm$0.9} \\
\midrule
\multicolumn{5}{c}{\textit{Order: T $\rightarrow$ I $\rightarrow$ A}} \\
\midrule
\texttt{BIAS\_IMAGE} & 60.3{\scriptsize\,$\pm$2.1} & 47.0{\scriptsize\,$\pm$2.2} & 37.8{\scriptsize\,$\pm$2.1} & 17.1{\scriptsize\,$\pm$1.7} \\
\texttt{BIAS\_AUDIO} & 0.9{\scriptsize\,$\pm$0.4} & 10.5{\scriptsize\,$\pm$1.3} & 0.3{\scriptsize\,$\pm$0.2} & 6.1{\scriptsize\,$\pm$1.0} \\
\texttt{BIAS\_TEXT} & 28.1{\scriptsize\,$\pm$2.0} & 31.6{\scriptsize\,$\pm$2.0} & 51.5{\scriptsize\,$\pm$2.2} & 62.0{\scriptsize\,$\pm$2.1} \\
\texttt{BIAS\_IMAGE\_AUDIO} & 0.1{\scriptsize\,$\pm$0.1} & 0.2{\scriptsize\,$\pm$0.2} & 0.0{\scriptsize\,$\pm$0.0} & 0.6{\scriptsize\,$\pm$0.3} \\
\texttt{BIAS\_IMAGE\_TEXT} & 0.5{\scriptsize\,$\pm$0.3} & 0.2{\scriptsize\,$\pm$0.2} & 2.4{\scriptsize\,$\pm$0.7} & 0.3{\scriptsize\,$\pm$0.2} \\
\texttt{BIAS\_AUDIO\_TEXT} & 0.0{\scriptsize\,$\pm$0.0} & 0.7{\scriptsize\,$\pm$0.4} & 0.0{\scriptsize\,$\pm$0.0} & 0.2{\scriptsize\,$\pm$0.2} \\
\texttt{HALLUCINATION} & 9.6{\scriptsize\,$\pm$1.3} & 9.1{\scriptsize\,$\pm$1.3} & 3.2{\scriptsize\,$\pm$0.8} & 5.8{\scriptsize\,$\pm$1.0} \\
\texttt{NO\_BIAS}& 0.6{\scriptsize\,$\pm$0.3} & 0.9{\scriptsize\,$\pm$0.4} & 4.9{\scriptsize\,$\pm$0.9} & 8.0{\scriptsize\,$\pm$1.2} \\
\midrule
\multicolumn{5}{c}{\textit{Order: T $\rightarrow$ A $\rightarrow$ I}} \\
\midrule
\texttt{BIAS\_IMAGE} & 65.5{\scriptsize\,$\pm$2.1} & 65.7{\scriptsize\,$\pm$2.1} & 43.5{\scriptsize\,$\pm$2.2} & 47.6{\scriptsize\,$\pm$2.2} \\
\texttt{BIAS\_AUDIO} & 1.0{\scriptsize\,$\pm$0.4} & 8.9{\scriptsize\,$\pm$1.2} & 0.4{\scriptsize\,$\pm$0.3} & 8.9{\scriptsize\,$\pm$1.2} \\
\texttt{BIAS\_TEXT} & 19.6{\scriptsize\,$\pm$1.7} & 14.5{\scriptsize\,$\pm$1.5} & 47.1{\scriptsize\,$\pm$2.2} & 28.9{\scriptsize\,$\pm$2.0} \\
\texttt{BIAS\_IMAGE\_AUDIO} & 0.1{\scriptsize\,$\pm$0.1} & 0.0{\scriptsize\,$\pm$0.0} & 0.0{\scriptsize\,$\pm$0.0} & 0.1{\scriptsize\,$\pm$0.1} \\
\texttt{BIAS\_IMAGE\_TEXT} & 2.6{\scriptsize\,$\pm$0.7} & 0.3{\scriptsize\,$\pm$0.2} & 2.9{\scriptsize\,$\pm$0.7} & 0.3{\scriptsize\,$\pm$0.2} \\
\texttt{BIAS\_AUDIO\_TEXT} & 0.0{\scriptsize\,$\pm$0.0} & 0.3{\scriptsize\,$\pm$0.2} & 0.0{\scriptsize\,$\pm$0.0} & 0.2{\scriptsize\,$\pm$0.2} \\
\texttt{HALLUCINATION} & 11.0{\scriptsize\,$\pm$1.4} & 10.0{\scriptsize\,$\pm$1.3} & 3.1{\scriptsize\,$\pm$0.8} & 9.9{\scriptsize\,$\pm$1.3} \\
\texttt{NO\_BIAS}& 0.3{\scriptsize\,$\pm$0.2} & 0.6{\scriptsize\,$\pm$0.3} & 3.1{\scriptsize\,$\pm$0.8} & 4.3{\scriptsize\,$\pm$0.9} \\
\bottomrule
\end{tabular}
}
\caption{Modality bias distribution (\%) for \textbf{Gemma4} across different evidence-type conditions and input orders. Modality input orders are denoted sequentially (e.g., $I \rightarrow A \rightarrow T$ represents Image followed by Audio, then Text).}
\label{tab:gemma_full_results}
\end{table}

%% file: tables/main_result_appendix.tex
\begin{table}[t]
\centering
\footnotesize
\resizebox{\columnwidth}{!}{
\begin{tabular}{l c c c c}
\toprule
\textbf{Bias Type} & \textbf{\makecell{Perc\textsubscript{I}-\\Perc\textsubscript{A}}} & \textbf{\makecell{Perc\textsubscript{I}-\\Prop\textsubscript{A}}} & \textbf{\makecell{Prop\textsubscript{I}-\\Perc\textsubscript{A}}} & \textbf{\makecell{Prop\textsubscript{I}-\\Prop\textsubscript{A}}} \\
\midrule
\texttt{BIAS\_IMAGE} & 73.4{\scriptsize\,$\pm$1.9} & 79.2{\scriptsize\,$\pm$1.8} & 40.1{\scriptsize\,$\pm$2.1} & 68.0{\scriptsize\,$\pm$2.0} \\
\texttt{BIAS\_AUDIO} & 4.8{\scriptsize\,$\pm$0.9} & 5.4{\scriptsize\,$\pm$1.0} & 17.2{\scriptsize\,$\pm$1.7} & 5.5{\scriptsize\,$\pm$1.0} \\
\texttt{BIAS\_TEXT} & 9.8{\scriptsize\,$\pm$1.3} & 8.0{\scriptsize\,$\pm$1.2} & 37.1{\scriptsize\,$\pm$2.1} & 22.5{\scriptsize\,$\pm$1.8} \\
\texttt{BIAS\_IMAGE\_AUDIO} & 2.5{\scriptsize\,$\pm$0.7} & 0.2{\scriptsize\,$\pm$0.2} & 0.1{\scriptsize\,$\pm$0.1} & 0.0{\scriptsize\,$\pm$0.0} \\
\texttt{BIAS\_IMAGE\_TEXT} & 0.3{\scriptsize\,$\pm$0.2} & 0.0{\scriptsize\,$\pm$0.0} & 0.6{\scriptsize\,$\pm$0.3} & 0.2{\scriptsize\,$\pm$0.2} \\
\texttt{BIAS\_AUDIO\_TEXT} & 0.0{\scriptsize\,$\pm$0.0} & 0.3{\scriptsize\,$\pm$0.2} & 0.1{\scriptsize\,$\pm$0.1} & 0.7{\scriptsize\,$\pm$0.4} \\
\texttt{HALLUCINATION} & 9.3{\scriptsize\,$\pm$1.3} & 7.0{\scriptsize\,$\pm$1.1} & 4.7{\scriptsize\,$\pm$0.9} & 1.7{\scriptsize\,$\pm$0.6} \\
\texttt{NO\_BIAS}& 0.1{\scriptsize\,$\pm$0.1} & 0.1{\scriptsize\,$\pm$0.1} & 0.3{\scriptsize\,$\pm$0.2} & 1.5{\scriptsize\,$\pm$0.5} \\
\bottomrule
\end{tabular}
}
\caption{Modality bias distribution (\%) for \textbf{Qwen3} across evidence-type conditions.}
\label{tab:qwen3_full_results}
\end{table}

\begin{table}[t]
\centering
\footnotesize
\resizebox{\columnwidth}{!}{
\begin{tabular}{l c c c c}
\toprule
\textbf{Bias Type} & \textbf{\makecell{Perc\textsubscript{I}-\\Perc\textsubscript{A}}} & \textbf{\makecell{Perc\textsubscript{I}-\\Prop\textsubscript{A}}} & \textbf{\makecell{Prop\textsubscript{I}-\\Perc\textsubscript{A}}} & \textbf{\makecell{Prop\textsubscript{I}-\\Prop\textsubscript{A}}} \\
\midrule
\texttt{BIAS\_IMAGE} & 59.1{\scriptsize\,$\pm$2.2} & 63.5{\scriptsize\,$\pm$2.1} & 47.1{\scriptsize\,$\pm$2.2} & 43.9{\scriptsize\,$\pm$2.2} \\
\texttt{BIAS\_AUDIO} & 5.9{\scriptsize\,$\pm$1.0} & 8.5{\scriptsize\,$\pm$1.2} & 12.3{\scriptsize\,$\pm$1.4} & 12.8{\scriptsize\,$\pm$1.5} \\
\texttt{BIAS\_TEXT} & 20.4{\scriptsize\,$\pm$1.8} & 11.6{\scriptsize\,$\pm$1.4} & 22.3{\scriptsize\,$\pm$1.8} & 19.0{\scriptsize\,$\pm$1.7} \\
\texttt{BIAS\_IMAGE\_AUDIO} & 0.3{\scriptsize\,$\pm$0.2} & 0.3{\scriptsize\,$\pm$0.2} & 0.0{\scriptsize\,$\pm$0.0} & 0.6{\scriptsize\,$\pm$0.3} \\
\texttt{BIAS\_IMAGE\_TEXT} & 0.1{\scriptsize\,$\pm$0.1} & 0.2{\scriptsize\,$\pm$0.2} & 0.6{\scriptsize\,$\pm$0.3} & 0.7{\scriptsize\,$\pm$0.4} \\
\texttt{BIAS\_AUDIO\_TEXT} & 0.1{\scriptsize\,$\pm$0.1} & 0.7{\scriptsize\,$\pm$0.4} & 0.1{\scriptsize\,$\pm$0.1} & 1.6{\scriptsize\,$\pm$0.5} \\
\texttt{HALLUCINATION} & 10.1{\scriptsize\,$\pm$1.3} & 8.8{\scriptsize\,$\pm$1.2} & 8.0{\scriptsize\,$\pm$1.2} & 8.1{\scriptsize\,$\pm$1.2} \\
\texttt{NO\_BIAS}& 4.3{\scriptsize\,$\pm$0.9} & 6.5{\scriptsize\,$\pm$1.1} & 9.8{\scriptsize\,$\pm$1.3} & 13.6{\scriptsize\,$\pm$1.5} \\
\bottomrule
\end{tabular}
}
\caption{Modality bias distribution (\%) for \textbf{MiniCPM} across evidence-type conditions.}
\label{tab:minicpm_full_results}
\end{table}

\begin{table}[t]
\centering
\footnotesize
\resizebox{\columnwidth}{!}{
\begin{tabular}{l c c c c}
\toprule
\textbf{Bias Type} & \textbf{\makecell{Perc\textsubscript{I}-\\Perc\textsubscript{A}}} & \textbf{\makecell{Perc\textsubscript{I}-\\Prop\textsubscript{A}}} & \textbf{\makecell{Prop\textsubscript{I}-\\Perc\textsubscript{A}}} & \textbf{\makecell{Prop\textsubscript{I}-\\Prop\textsubscript{A}}} \\
\midrule
\texttt{BIAS\_IMAGE} & 68.9{\scriptsize\,$\pm$2.0} & 85.1{\scriptsize\,$\pm$1.6} & 37.6{\scriptsize\,$\pm$2.1} & 73.6{\scriptsize\,$\pm$1.9} \\
\texttt{BIAS\_AUDIO} & 4.2{\scriptsize\,$\pm$0.9} & 2.0{\scriptsize\,$\pm$0.6} & 17.2{\scriptsize\,$\pm$1.7} & 1.1{\scriptsize\,$\pm$0.5} \\
\texttt{BIAS\_TEXT} & 9.2{\scriptsize\,$\pm$1.3} & 2.7{\scriptsize\,$\pm$0.7} & 33.6{\scriptsize\,$\pm$2.1} & 4.6{\scriptsize\,$\pm$0.9} \\
\texttt{BIAS\_IMAGE\_AUDIO} & 7.3{\scriptsize\,$\pm$1.1} & 0.7{\scriptsize\,$\pm$0.4} & 1.4{\scriptsize\,$\pm$0.5} & 2.3{\scriptsize\,$\pm$0.7} \\
\texttt{BIAS\_IMAGE\_TEXT} & 3.2{\scriptsize\,$\pm$0.8} & 0.6{\scriptsize\,$\pm$0.3} & 3.7{\scriptsize\,$\pm$0.8} & 0.7{\scriptsize\,$\pm$0.4} \\
\texttt{BIAS\_AUDIO\_TEXT} & 0.3{\scriptsize\,$\pm$0.2} & 0.7{\scriptsize\,$\pm$0.4} & 0.4{\scriptsize\,$\pm$0.3} & 0.1{\scriptsize\,$\pm$0.1} \\
\texttt{HALLUCINATION} & 6.3{\scriptsize\,$\pm$1.1} & 8.0{\scriptsize\,$\pm$1.2} & 4.6{\scriptsize\,$\pm$0.9} & 6.3{\scriptsize\,$\pm$1.1} \\
\texttt{NO\_BIAS}& 0.7{\scriptsize\,$\pm$0.4} & 0.3{\scriptsize\,$\pm$0.2} & 1.6{\scriptsize\,$\pm$0.5} & 11.6{\scriptsize\,$\pm$1.4} \\
\bottomrule
\end{tabular}
}
\caption{Modality bias distribution (\%) for \textbf{Gemini3} across evidence-type conditions.}
\label{tab:gemini3_full_results}
\end{table}

%% file: tables/CD_full_results.tex
\begin{table}[t]
\centering
\setlength{\tabcolsep}{3pt}
\footnotesize
\resizebox{\linewidth}{!}{
\begin{tabular}{llcccc}
\toprule
\textbf{Bias Type} & \textbf{Method} & \textbf{\makecell{Perc\textsubscript{I}-\\Perc\textsubscript{A}}} & \textbf{\makecell{Perc\textsubscript{I}-\\Prop\textsubscript{A}}} & \textbf{\makecell{Prop\textsubscript{I}-\\Perc\textsubscript{A}}} & \textbf{\makecell{Prop\textsubscript{I}-\\Prop\textsubscript{A}}} \\
\midrule
\multirow{2}{*}{\texttt{BIAS\_IMAGE}} & Baseline & 60.30{\scriptsize\,$\pm$2.1} & 52.4{\scriptsize\,$\pm$2.2} & 74.4{\scriptsize\,$\pm$1.9} & 74.9{\scriptsize\,$\pm$1.9} \\
 & CD & 47.6{\scriptsize\,$\pm$2.2} & 31.3{\scriptsize\,$\pm$2.0} & 39.6{\scriptsize\,$\pm$2.1} & 38.8{\scriptsize\,$\pm$2.1} \\
\midrule
\multirow{2}{*}{\texttt{BIAS\_AUDIO}} & Baseline & 0.9{\scriptsize\,$\pm$0.4} & 25.4{\scriptsize\,$\pm$1.9} & 2.7{\scriptsize\,$\pm$0.7} & 5.4{\scriptsize\,$\pm$1.0} \\
 & CD & 1.1{\scriptsize\,$\pm$0.5} & 16.6{\scriptsize\,$\pm$1.6} & 1.3{\scriptsize\,$\pm$0.5} & 5.5{\scriptsize\,$\pm$1.0} \\
\midrule
\multirow{2}{*}{\texttt{BIAS\_TEXT}} & Baseline & 25.1{\scriptsize\,$\pm$1.9} & 9.0{\scriptsize\,$\pm$1.3} & 8.7{\scriptsize\,$\pm$1.2} & 5.1{\scriptsize\,$\pm$1.0} \\
 & CD & 34.0{\scriptsize\,$\pm$2.1} & 22.5{\scriptsize\,$\pm$1.8} & 12.4{\scriptsize\,$\pm$1.4} & 7.3{\scriptsize\,$\pm$1.1} \\
\midrule
\multirow{2}{*}{\makecell[l]{\texttt{BIAS\_IMAGE\_}\\\texttt{AUDIO}}} & Baseline & 0.1{\scriptsize\,$\pm$0.1} & 0.2{\scriptsize\,$\pm$0.2} & 0.2{\scriptsize\,$\pm$0.2} & 0.3{\scriptsize\,$\pm$0.2} \\
 & CD & 0.1{\scriptsize\,$\pm$0.1} & 0.3{\scriptsize\,$\pm$0.2} & 0.1{\scriptsize\,$\pm$0.1} & 0.6{\scriptsize\,$\pm$0.3} \\
\midrule
\multirow{2}{*}{\makecell[l]{\texttt{BIAS\_IMAGE\_}\\\texttt{TEXT}}} & Baseline & 0.2{\scriptsize\,$\pm$0.2} & 0.1{\scriptsize\,$\pm$0.1} & 1.4{\scriptsize\,$\pm$0.5} & 0.3{\scriptsize\,$\pm$0.2} \\
 & CD & 0.2{\scriptsize\,$\pm$0.2} & 0.1{\scriptsize\,$\pm$0.1} & 13.6{\scriptsize\,$\pm$1.5} & 4.0{\scriptsize\,$\pm$0.9} \\
\midrule
\multirow{2}{*}{\makecell[l]{\texttt{BIAS\_AUDIO\_}\\\texttt{TEXT}}} & Baseline & 0.0{\scriptsize\,$\pm$0.0} & 1.0{\scriptsize\,$\pm$0.4} & 0.0{\scriptsize\,$\pm$0.0} & 0.2{\scriptsize\,$\pm$0.2} \\
 & CD & 0.1{\scriptsize\,$\pm$0.1} & 5.9{\scriptsize\,$\pm$1.0} & 0.1{\scriptsize\,$\pm$0.1} & 0.7{\scriptsize\,$\pm$0.4} \\
\midrule
\multirow{2}{*}{\texttt{HALLUCINATION}} & Baseline & 11.6{\scriptsize\,$\pm$1.4} & 10.5{\scriptsize\,$\pm$1.3} & 7.3{\scriptsize\,$\pm$1.1} & 6.9{\scriptsize\,$\pm$1.1} \\
 & CD & 12.1{\scriptsize\,$\pm$1.4} & 5.9{\scriptsize\,$\pm$1.0} & 5.5{\scriptsize\,$\pm$1.0} & 4.7{\scriptsize\,$\pm$0.9} \\
\midrule
\multirow{2}{*}{\texttt{NO\_BIAS}} & Baseline & 2.0{\scriptsize\,$\pm$0.6} & 1.7{\scriptsize\,$\pm$0.6} & 5.5{\scriptsize\,$\pm$1.0} & 7.2{\scriptsize\,$\pm$1.1} \\
 & CD & 5.0{\scriptsize\,$\pm$1.0} & 17.6{\scriptsize\,$\pm$1.7} & 27.7{\scriptsize\,$\pm$2.0} & 38.6{\scriptsize\,$\pm$2.1} \\
\bottomrule
\end{tabular}
}
\caption{Detailed bias-type distribution (\%) for the baseline and CD (Contrastive Decoding) on Gemma~4 across evidence-type conditions.}
\label{tab:cd_full_results}
\end{table}

%% file: tables/omnibench_cd.tex
\begin{table}[t!]
\centering
\small
\setlength{\tabcolsep}{4pt}
\renewcommand{\arraystretch}{1.12}
\resizebox{\columnwidth}{!}{
\begin{tabular}{@{}l@{\hspace{6pt}}r@{\hspace{6pt}}r@{\hspace{6pt}}r@{}}
\toprule
\textbf{Split} & \textbf{$N$} & \textbf{Baseline (\%)} & \textbf{CD (\%)} \\
\midrule
\multicolumn{4}{@{}l}{\textit{Overall}} \\
Overall & 1142 & 38.4 & 37.4 \\
\midrule
\multicolumn{4}{@{}l}{\textit{By audio type}} \\
Speech & 771 & 41.1 & 40.1 \\
Sound event & 265 & 35.1 & 33.6 \\
Music & 106 & 27.4 & 27.4 \\
\midrule
\multicolumn{4}{@{}l}{\textit{By task type}} \\
Action and Activity & 251 & 37.5 & 33.5 \\
Plot Inference & 237 & 31.2 & 35.0 \\
Story Description & 230 & 33.9 & 36.5 \\
\parbox[t]{0.34\columnwidth}{\raggedright Object Identification\\and Description} &
\parbox[t]{1.6em}{\raggedleft 211} &
\parbox[t]{2.4em}{\raggedleft 49.3} &
\parbox[t]{2.4em}{\raggedleft 45.0} \\
\parbox[t]{0.34\columnwidth}{\raggedright Contextual and Environmental\\Questions} &
\parbox[t]{1.6em}{\raggedleft 141} &
\parbox[t]{2.4em}{\raggedleft 51.8} &
\parbox[t]{2.4em}{\raggedleft 45.4} \\
Identity and Relationship & 32 & 21.9 & 31.2 \\
Text and Symbols & 25 & 20.0 & 12.0 \\
Count and Quantity & 15 & 26.7 & 26.7 \\
\bottomrule
\end{tabular}
}
\caption{\textsc{OmniBench} multiple-choice accuracy (\%) for Gemma~4 under baseline decoding and contrastive decoding (CD). Invalid parses ($N/A$) are counted as incorrect.}
\label{tab:omnibench-cd}
\end{table}